\documentclass[review,5p,twocolumn]{elsarticle}

\usepackage{lineno,hyperref}
\modulolinenumbers[5]
\graphicspath{{figures/}}
\usepackage{caption}
\usepackage{setspace}
\usepackage{textcomp}
\usepackage{bbding}
\usepackage{booktabs}
\usepackage{threeparttable}
\usepackage{epsfig}
\usepackage{subfigure}
\usepackage{graphicx}
\usepackage{algorithm}
\usepackage{algorithmic}
\usepackage{amssymb}
\usepackage{mathrsfs} 
\usepackage{amsmath,bm}
\usepackage{array,multirow}
\usepackage{mdwlist}
\usepackage{paralist}
\usepackage{url}
\usepackage{graphicx}
\usepackage{epstopdf}
\usepackage{times} 
\usepackage{helvet} 
\usepackage{courier} 
\usepackage{url} 
\usepackage{graphicx} 
\usepackage{xcolor}

\usepackage{breakurl}
\usepackage{amssymb}
\usepackage{makecell}
\usepackage{multirow}
\usepackage{rotating}
\usepackage{array}
\usepackage{times}
\usepackage{graphicx}
 \usepackage{psfrag}
\usepackage{subfigure}
\usepackage{rotating}
\usepackage{url}
\usepackage{breakurl}
\usepackage{rotating, graphicx}
\usepackage{amsmath}
\usepackage{booktabs}
\usepackage{lscape}
\usepackage{longtable}
\usepackage{threeparttable}
\usepackage{bm}

\usepackage{times}
\usepackage{epsfig}
\usepackage{graphicx}
\usepackage{amsmath}
\usepackage{amssymb}

\graphicspath{{figures/}}

\begin{document}

\begin{frontmatter}

\title{Recent Advances in Deep Learning for Object Detection}

\author[mymainaddress]{Xiongwei Wu\corref{mycorrespondingauthor}}
\cortext[mycorrespondingauthor]{Corresponding author}
\ead{xwwu.2015@phdis.smu.edu.sg}

\author[mymainaddress]{Doyen Sahoo}
\ead{doyens.smu.edu.sg}

\author[mymainaddress,mysecondaryaddress]{Steven C.H. Hoi}
\ead{chhoi@smu.edu.sg}

\address[mymainaddress]{School of Information System, Singapore Management University}
\address[mysecondaryaddress]{Salesforce Research Asia}

\begin{abstract}
Object detection is a fundamental visual recognition problem in computer vision and has been widely studied in the past decades. Visual object detection aims to find objects of certain target classes with precise localization in a given image and assign each object instance a corresponding class label. Due to the tremendous successes of deep learning based image classification, object detection techniques using deep learning have been actively studied in recent years. In this paper, we give a comprehensive survey of recent advances in visual object detection with deep learning. By reviewing a large body of recent related work in literature, we systematically analyze the existing object detection frameworks and organize the survey into three major parts: (i) detection components, (ii) learning strategies, and (iii) applications \& benchmarks. In the survey, we cover a variety of factors affecting the detection performance in detail, such as detector architectures, feature learning, proposal generation, sampling strategies, etc. Finally, we discuss several future directions to facilitate and spur future research for visual object detection with deep learning. 
\end{abstract}


\end{frontmatter}


\begin{figure}[ht!]
\centering

\subfigure[Image Classification] {
		\includegraphics[width=0.23\textwidth]{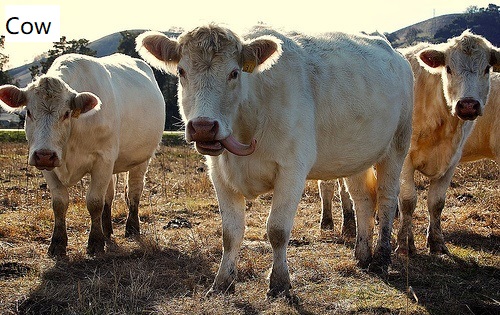}}
\subfigure[Object Detection]{
		\includegraphics[width=0.23\textwidth]{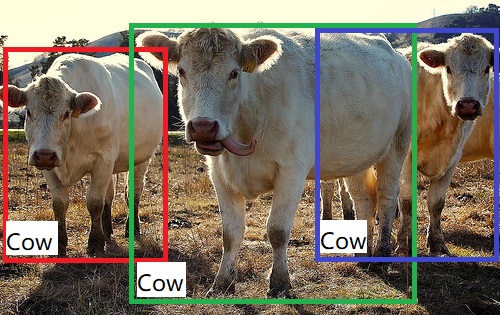}}
\subfigure[Semantic Segmentation] {
		\includegraphics[width=0.23\textwidth]{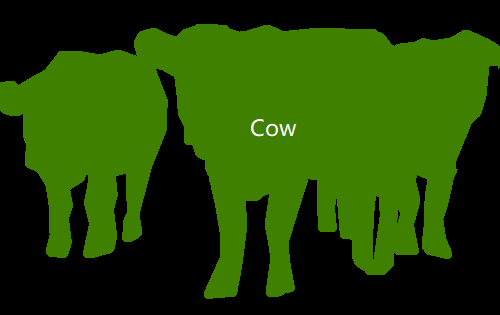}}
\subfigure[Instance Segmentation]{
		\includegraphics[width=0.23\textwidth]{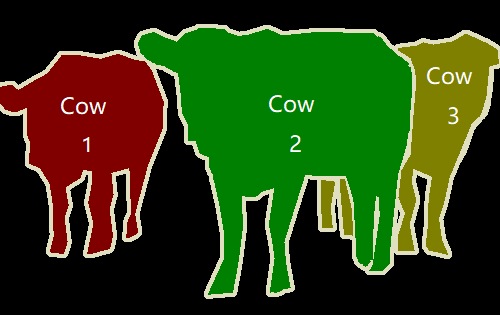}}
	\caption{Comparison of different visual recognition tasks in computer vision. (a) ``Image Classification" only needs to assign categorical class labels to the image; (b) ``Object detection" not only predict categorical labels but also localize each object instance via bounding boxes; (c) ``Semantic segmentation" aims to predict categorical labels for each pixel, without differentiating object instances; (d) ``Instance segmentation", a special setting of object detection, differentiates different object instances by pixel-level segmentation masks. } \label{fig:clsvsdet}
\end{figure}

\section{Introduction}\label{sec:introduction}
In the field of computer vision, there are several fundamental visual recognition problems: image classification~\cite{he2016deep}, object detection and instance segmentation~\cite{girshick2014rich,he2017mask}, and semantic segmentation~\cite{chen2014semantic} (see Fig. \ref{fig:clsvsdet}). In particular, image classification (Fig 1.\ref{fig:clsvsdet}(a)), aims to recognize semantic categories of objects in a given image. Object detection not only recognizes object categories, but also predicts the location of each object by a bounding box (Fig. \ref{fig:clsvsdet}(b)). Semantic segmentation (Fig. \ref{fig:clsvsdet}(c)) aims to predict pixel-wise classifiers to assign a specific category label to each pixel, thus providing an even richer understanding of an image. However, in contrast to object detection, semantic segmentation does not distinguish between multiple objects of the same category. A relatively new setting at the intersection of object detection and semantic segmentation, named ``instance segmentation" (Fig. \ref{fig:clsvsdet}(d)), is proposed to identify different objects and assign each of them a separate categorical pixel-level mask. In fact, instance segmentation can be viewed as a special setting of object detection, where instead of localizing an object by a bounding box, pixel-level localization is desired. In this survey, we direct our attention to review the major efforts in deep learning based object detection. A good detection algorithm should have a strong understanding of semantic cues as well as the spatial information about the image. In fact, object detection is the basic step towards many computer vision applications, such as face recognition ~\cite{sun2015deepid3,sun2014deep,liu2017sphereface}, pedestrian detection~\cite{li2018scale,hosang2015taking,angelova2015real}, video analysis~\cite{karpathy2014large,mobahi2009deep}, and logo detection~\cite{hoi2015logo,su2017deep,su2018scalable}.

In the early stages, before the deep learning era, the pipeline of object detection was divided into three steps: i) proposal generation; ii) feature vector extraction; and iii) region classification. During proposal generation, the objective was to search locations in the image which may contain objects. These locations are also called regions of interest (roi). An intuitive idea is to scan the whole image with sliding windows ~\cite{vedaldi2009multiple,viola2001rapid,harzallah2009combining,dalal2005histograms,viola2004robust}. In order to capture information about multi-scale and different aspect ratios of objects, input images were resized into different scales and multi-scale windows were used to slide through these images. During the second step, on each location of the image, a fixed-length feature vector was obtained from the sliding window, to capture discriminative semantic information of the region covered. This feature vector was commonly encoded by low-level visual descriptors such as SIFT(Scale Invariant Feature Transform)~\cite{lowe1999object}, Haar~\cite{lienhart2002extended}, HOG(Histogram of Gradients)~\cite{dalal2005histograms} or SURF(Speeded Up Robust Features)~\cite{bay2006surf}, which showed a certain robustness to scale, illumination and rotation variance. Finally, in the third step, the region classifiers were learned to assign categorical labels to the covered regions. Commonly, support vector machines(SVM)~\cite{hearst1998support} were used here due to their good performance on small scale training data. In addition, some classification techniques such as bagging~\cite{opitz1999popular}, cascade learning~\cite{viola2004robust} and adaboost~\cite{freund1996experiments} were used in region classification step, leading to further improvements in detection accuracy.

\begin{figure*}[htb]
\centering
\includegraphics[width=7.5in]{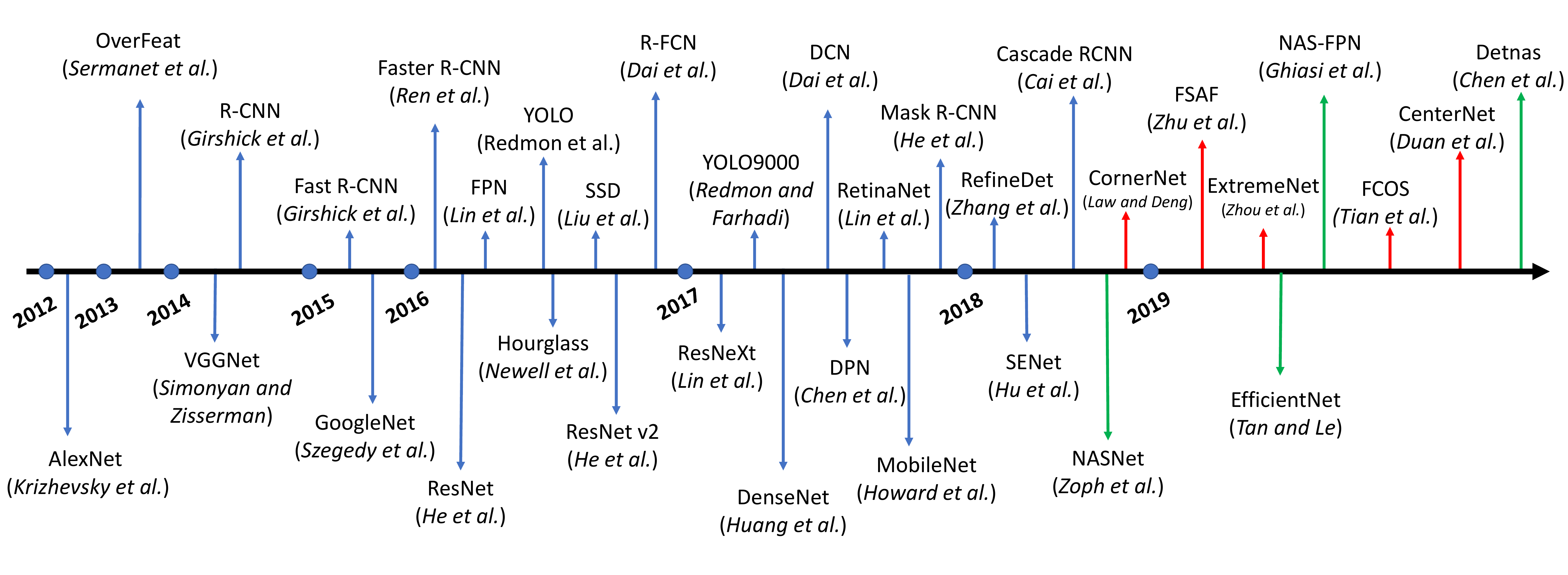}
\caption{Major milestone in object detection research based on deep convolution neural networks since 2012. The trend in the last year has been designing object detectors based on anchor-free(in red) and AutoML(in green) techniques, which are potentially two important research directions in the future.}
\label{fig:milestone}
\end{figure*}

Most of the successful traditional methods for object detection focused on carefully designing feature descriptors to obtain embedding for a region of interest. With the help of good feature representations as well as robust region classifiers, impressive results~\cite{yu2010object,felzenszwalb2008discriminatively} were achieved on Pascal VOC dataset~\cite{Everingham10} (a publicly available dataset used for benchmarking object detection). Notably, deformable part based machines (DPMs)~\cite{felzenszwalb2010object}, a breakthrough detection algorithm, were 3-time winners on VOC challenges in 2007, 2008 and 2009. DPMs learn and integrate multiple part models with a deformable loss and mine hard negative examples with a latent SVM for discriminative training. However, during 2008 to 2012, the progress on Pascal VOC based on these traditional methods had become incremental, with minor gains from building complicated ensemble systems. This showed the limitations of these traditional detectors. Most prominently, these limitations included: (i) during proposal generation, a huge number of proposals were generated, and many of them were redundant; this resulted in a large number of false positives during classification. Moreover, window scales were designed manually and heuristically, and could not match the objects well; (ii) feature descriptors were hand-crafted based on low level visual cues~\cite{lowe2004distinctive,ojala2002multiresolution,bay2006surf}, which made it difficult to capture representative semantic information in complex contexts. (iii) each step of the detection pipeline was designed and optimized separately, and thus could not obtain a global optimal solution for the whole system.

After the success of applying deep convolutional neural networks(DCNN) for image classification~\cite{krizhevsky2012imagenet,he2016deep}, object detection also achieved remarkable progress based on deep learning techniques~\cite{ren2015faster,girshick2014rich}. The new deep learning based algorithms outperformed the traditional detection algorithms by huge margins. Deep convolutional neural network is a biologically-inspired structure for computing hierarchical features. An early attempt to build such a hierarchical and spatial-invariant model for image classification was ``neocognitron''~\cite{fukushima1982neocognitron} proposed by Fukushima. However, this early attempt lacked effective optimization techniques for supervised learning. Based on this model, Lecun et al.~\cite{lecun1998gradient} optimized a convolutional neural network by stochastic gradient descent (SGD) via back-propagation and showed competitive performance on digit recognition. After that, however, deep convolutional neural networks were not heavily explored, with support vector machines becoming more prominent. This was because deep learning had some limitations: (i) lack of large scale annotated training data, which caused overfitting; (ii) limited computation resources; and (iii) weak theoretical support compared to SVMs. In 2009, Jia et al.~\cite{deng2009imagenet} collected a large scale annotated image dataset ImageNet which contained 1.2M high resolution images, making it possible to train deep models with large scale training data. With the development of computing resources on parallel computing systems(such as GPU clusters), in 2012 Krizhevsky et al.~\cite{krizhevsky2012imagenet} trained a large deep convolutional model with ImageNet dataset and showed significant improvement on Large Scale Visual Recognition Challenge(ILSVRC) compared to all other approaches. After the success of applying DCNN for classification, deep learning techniques were quickly adapted to other vision tasks and showed promising results compared to the traditional methods. 

In contrast to hand-crafted descriptors used in traditional detectors, deep convolutional neural networks generate hierarchical feature representations from raw pixels to high level semantic information, which is learned automatically from the training data and shows more discriminative expression capability in complex contexts. Furthermore, benefiting from the powerful learning capacity, a deep convolutional neural network can obtain a better feature representation with a larger dataset, while the learning capacity of traditional visual descriptors are fixed, and can not improve when more data becomes available. These properties made it possible to design object detection algorithms based on deep convolutional neural networks which could be optimized in an end-to-end manner, with more powerful feature representation capability.

\begin{figure*}[htb]
\centering
\includegraphics[width=7.2in]{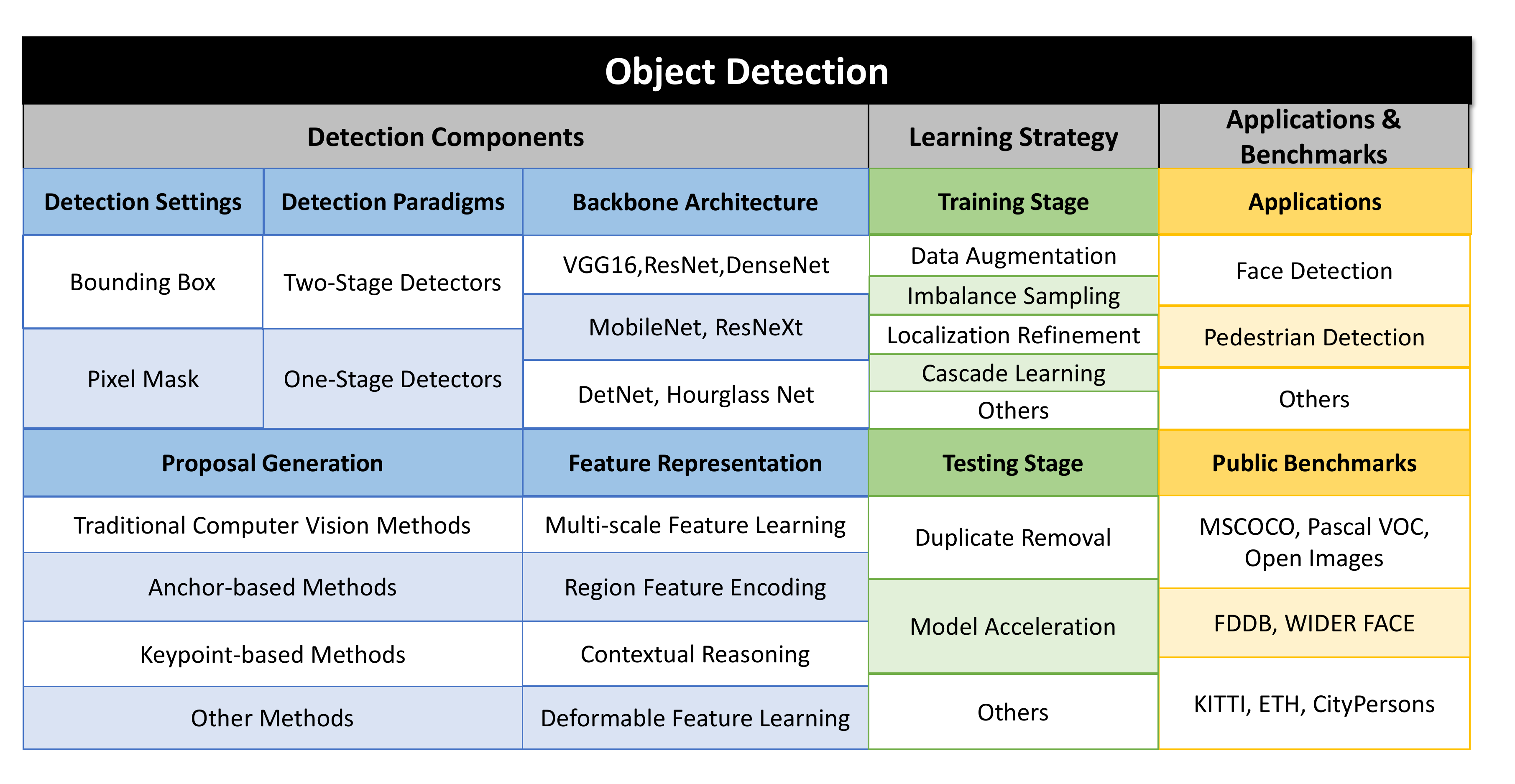}\vspace{-2mm}
\caption{Taxonomy of key methodologies in this survey. We categorize various contributions for deep learning based object detection into three major categories: Detection Components, Learning Strategies, Applications and Benchmarks. We review each of these categories in detail.}
\label{fig:pipeline}
\end{figure*}

Currently, deep learning based object detection frameworks can be primarily divided into two families: (i) two-stage detectors, such as Region-based CNN (R-CNN)~\cite{girshick2014rich} and its variants ~\cite{girshick2015fast,ren2015faster,lin2016fpn} and (ii) one-stage detectors, such as YOLO~\cite{redmon2016you} and its variants~\cite{redmon2017yolo9000,liu2016ssd}. Two-stage detectors first use a proposal generator to generate a sparse set of proposals and extract features from each proposal, followed by region classifiers which predict the category of the proposed region. One-stage detectors directly make categorical prediction of objects on each location of the feature maps without the cascaded region classification step. Two-stage detectors commonly achieve better detection performance and report state-of-the-art results on public benchmarks, while one-stage detectors are significantly more time-efficient and have greater applicability to real-time object detection. Figure \ref{fig:milestone} also illustrates the major developments and milestones of deep learning based object detection techniques after 2012. We will cover basic ideas of these key techniques and analyze them in a systematic manner in the survey.

The goal of this survey is to present a comprehensive understanding of deep learning based object detection algorithms. Fig. \ref{fig:pipeline} shows a taxonomy of key methodologies to be covered in this survey.  We review various contributions in deep learning based object detection and categorize them into three groups: detection components, learning strategies, and applications \& benchmarks. For detection components, we first introduce two detection settings: bounding box level(bbox-level) and pixel mask level(mask-level) localization. Bbox-level algorithms require to localize objects by rectangle bounding boxes, while more precise pixel-wise masks are required to segment objects in mask-level algorithms. Next, we summarize the representative frameworks of two detection families: two-stage detection and one-stage detection. Then we give a detailed survey of each detection component, including backbone architecture, proposal generation and feature learning. For learning strategies, we first highlight the importance of learning strategy of detection due to the difficulty of training detectors, and then introduce the optimization techniques for both training and testing stages in detail. Finally, we review some real-world object detection based applications including face detection, pedestrian detection, logo detection and video analysis. We also discuss publicly available and commonly used benchmarks and evaluation metrics for these detection tasks. Finally we show the state-of-the-art results of generic detection on public benchmarks over the recent years. 

We hope our survey can provide a timely review for researchers and practitioners to further catalyze research on detection systems. The rest of the paper are organized as follows: in Section \ref{sec:problem}, we give a standard problem setting of object detection. The details of detector components are listed in Section \ref{sec:detcomp}. Then the learning strategies are presented in Section \ref{sec:lrstr}. Detection algorithms for real-world applications and benchmarks are provided in Section \ref{sec:app} and Section \ref{sec:bench}. State-of-the-art results of generic detection are listed in Section \ref{sec:sota}. Finally, we conclude and discuss future directions in Section \ref{sec:conc}.

\section{Problem Settings}\label{sec:problem}

In this section, we present the formal problem setting for object detection based on deep learning. Object detection involves both recognition (e.g., ``object classification") and localization (e.g., ``location regression") tasks. An object detector needs to distinguish objects of certain target classes from backgrounds in the image with precise localization and correct categorical label prediction to each object instance. Bounding boxes or pixel masks are predicted to localize these target object instances. 

More formally, assume we are given a collection of $N$ annotated images $\begin{Bmatrix} x_1,x_2,...,x_N \end{Bmatrix}$, and for $i^{th}$ image $x_i$, there are $M_i$ objects belonging to $C$ categories with annotations:
\begin{align}
y_i = \begin{Bmatrix} (c_1^i, b_1^i), (c_2^i, b_2^i), ...,(c_{M_{i}}^i, b_{M_{i}}^i)
\end{Bmatrix}
\end{align}
where $c_j^i$( $c_j^i \in C$) and $b_j^i$ (bounding box or pixel mask of the object) denote categorical and spatial labels of j-th object in $x_i$ respectively. The detector is $f$
parameterized by $\theta$. For $x_i$, the prediction $y_{\text{pred}}^i$ shares the same format as $y_i$:
\begin{align}
y_{\text{pred}}^i = \begin{Bmatrix} (c_{\text{pred}_1}^i, b_{\text{pred}_1}^i), (c_{\text{pred}_2}^i, b_{\text{pred}_2}^i), ...)
\end{Bmatrix}
\end{align}
Finally a loss function $\ell$ is set to optimize detector as:
\begin{align}
  \ell(x, \theta) = \frac{1}{N}\sum_{i=1}^N \ell(y_\text{pred}^i, x_i, y_i; \theta) + \frac{\lambda}{2} \left \| \theta \right \|^2_2
\end{align}
where the second term is a regularizer, with trade-off parameter $\lambda$. Different loss functions such as softmax loss~\cite{girshick2015fast} and focal loss~\cite{lin2017focal} impact the final detection performance, and we will discuss these functions in Section \ref{sec:lrstr}.

At the time of evaluation, a metric called intersection-over-union (IoU) between objects and predictions is used to evaluate the quality of localization(we omit index $i$ here):
\begin{align}
\text{IoU}(b_\text{pred}, b_\text{gt}) =\frac{\text{Area}(b_\text{pred} \bigcap b_\text{gt})}{\text{Area}(b_\text{pred} \bigcup b_\text{gt})}
\end{align}
Here, $b_\text{gt}$ refers to the ground truth bbox or mask. An IoU threshold $\Omega$ is set to determine whether a prediction {\it tightly} covers the object or not(i.e. IoU $ \ge \Omega$; commonly researchers set $\Omega =0.5$). For object detection, a prediction with correct categorical label as well as successful localization prediction (meeting the IoU criteria) is considered as positive, otherwise it's a negative prediction:
\begin{align}
\text{Prediction} = \left\{ \begin{matrix}
 \text{Positive} & c_\text{pred}=c_\text{gt} \text{ } \text{and} \text{ } \text{IoU}(b_{\text{pred}}, b_{\text{gt}}) > \Omega\\
 \text{Negative}& \text{otherwise}
\end{matrix}\right.\end{align}
For generic object detection problem evaluation, mean average precision(mAP) over $C$ classes is used for evaluation, and in real world scenarios such as pedestrian detection, different evaluation metrics are used which will be discussed in Section \ref{sec:app}.
In addition to detection accuracy, inference speed is also an important metric to evaluate object detection algorithms. Specifically, if we wish to detect objects in a video stream (real-time detection), it is imperative to have a detector that can process this information quickly. Thus, the detector efficiency is also evaluated on Frame per second (FPS), i.e., how many images it can process per second. Commonly a detector that can achieve an inference speed of 20 FPS, is considered to be a real-time detector.

\section{Detection Components}\label{sec:detcomp}

In this section, we introduce different components of object detection. The first is about the choice of object detection paradigm. We first introduce the concepts of two detection settings: bbox-level and mask-level algorithms. Then, We introduce two major object detection paradigms: two-stage detectors and one-stage detectors. Under these paradigms, detectors can use a variety of deep learning backbone architectures, proposal generators, and feature representation modules. 

\subsection{Detection Settings}
There are two settings in object detection: i) vanilla object detection (bbox-level localization) and ii) instance segmentation (pixel-level or mask-level localization). Vanilla object detection has been more extensively studied and is considered as the traditional detection setting, where the goal is to localize objects by rectangle bounding boxes. In vanilla object detection algorithms, only bbox annotations are required, and in evaluation, the IoU between predicted bounding box with the ground truth is calculated to measure the performance. Instance segmentation is a relatively new setting and is based on traditional detection setting. Instance segmentation requires to segment each object by a pixel-wise mask instead of a rough rectangle bounding box. Due to more precise pixel-level prediction, instance segmentation is more sensitive to spatial misalignment, and thus has higher requirement to process the spatial information. The evaluation metric of instance segmentation is almost identical to the bbox-level detection, except that the IoU computation is performed on mask predictions. Though the two detection settings are slightly different, the main components introduced later can mostly be shared by the two settings. 

\subsection{Detection Paradigms}
Current state-of-the-art object detectors with deep learning can be mainly divided into two major categories: two-stage detectors and one-stage detectors.  For a two-stage detector, in the first stage, a sparse set of proposals is generated; and in the second stage, the feature vectors of generated proposals are encoded by deep convolutional neural networks followed by making the object class predictions. An one-stage detector does not have a separate stage for proposal generation (or learning a proposal generation). They typically consider all positions on the image as potential objects, and try to classify each region of interest as either background or a target object. Two-stage detectors often reported state-of-the-art results on many public benchmark datasets. However, they generally fall short in terms of lower inference speeds. One-stage detectors are much faster and more desired for real-time object detection applications, but have a relatively poor performance compared to the two-stage detectors.

\subsubsection{Two-stage Detectors}
Two-stage detectors split the detection task into two stages: (i) proposal generation; and (ii) making predictions for these proposals. During the proposal generation phase, the detector will try to identify regions in the image which may potentially be objects. 
The idea is to propose regions with a high recall, such that all objects in the image belong to at least one of these proposed region. In the second stage, a deep-learning based model is used to classify these proposals with the right categorical labels. The region may either be a background, or an object from one of the predefined class labels . Additionally, the model may refine the original localization suggested by the proposal generator. Next, we review some of the most influential efforts among two-stage detectors. 

\textbf{R-CNN}~\cite{girshick2014rich} is a pioneering two-stage object detector proposed by Girshick et al. in 2014. Compared to the previous state-of-the-art methods based on a traditional detection framework SegDPM~\cite{fidler2013bottom} with 40.4\% mAP on Pascal VOC2010, R-CNN significantly improved the detection performance and obtained 53.7\% mAP. The pipeline of R-CNN can be divided into three components: i) proposal generation, ii) feature extraction and iii) region classification. For each image, R-CNN generates a sparse set of proposals (around 2,000 proposals) via Selective Search~\cite{uijlings2013selective}, which is designed to reject regions that can easily be identified as background regions. Then, each proposal is cropped and resized into a fixed-size region and is encoded into a (e.g. 4,096 dimensional) feature vector by a deep convolutional neural network, followed by a one-vs-all SVM classifier. Finally the bounding box regressors are learned using the extracted features as input in order to make the original proposals tightly bound the objects. Compared to traditional hand-crafted feature descriptors, deep neural networks generate hierarchical features and capture different scale information in different layers, and finally produce robust and discriminative features for classification. utilize the power of transfer learning, R-CNN adopts weights of convolutional networks pre-trained on ImageNet. The last fully connected layer (FC layer) is re-initialized for the detection task. The whole detector is then finetuned on the pre-trained model. This transfer of knowledge from the Imagenet dataset offers significant performance gains. In addition, R-CNN rejects huge number of easy negatives before training, which helps improve learning speed and reduce false positives.

However, R-CNN faces some critical shortcomings: i) the features of each proposal were extracted by deep convolutional networks \textit{separately} (i.e., computation was not shared), which led to heavily duplicated computations. Thus, R-CNN was extremely time-consuming for training and testing; ii) the three steps of R-CNN (proposal generation, feature extraction and region classification) were independent components and the whole detection framework could not be optimized in an end-to-end manner, making it difficult to obtain global optimal solution; and iii) Selective Search relied on low-level visual cues and thus struggled to generate high quality proposals in complex contexts. Moreover, it is unable to enjoy the benefits of GPU acceleration.

Inspired by the idea of spatial pyramid matching (SPM)~\cite{kleban2008spatial}, He et al. proposed \textbf{SPP-net}~\cite{he2014spatial} to accelerate R-CNN as well as learn more discriminative features. Instead of cropping proposal regions and feeding into CNN model separately, SPP-net computes the feature map from the whole image using a deep convolutional network and extracts fixed-length feature vectors on the feature map by a Spatial Pyramid Pooling (SPP) layer. SPP partitions the feature map into an $N\times N$ grid, for multiple values of $N$ (thus allowing obtaining information at different scales), and performs pooling on each cell of the grid, to give a feature vector. The feature vectors obtained from each $N\times N$ grid are concatenated to give the representation for the region. The extracted features are fed into region SVM classifiers and bounding box regressors. In contrast to RCNN, SPP-layer can also work on images/regions at various scales and aspect ratios without resizing them. Thus, it does not suffer from information loss and unwanted geometric distortion. 

SPP-net achieved better results and had a significantly faster inference speed compared to R-CNN. However, the training of SPP-net was still multi-stage and thus it could not be optimized end-to-end (and required extra cache memory to store extracted features). In addition, SPP layer did not back-propagate gradients to convolutional kernels and thus all the parameters before the SPP layer were frozen. This significantly limited the learning capability of deep backbone architectures. Girshick et al. proposed ~\textbf{Fast R-CNN}~\cite{girshick2015fast}, a multi-task learning detector which addressed these two limitations of SPP-net. Fast R-CNN (like SPP-Net) also computed a feature map for the whole image and extracted fixed-length region features on the feature map. Different from SPP-net, Fast R-CNN used ROI Pooling layer to extract region features. \textit{{ ROI pooling}} layer is a special case of SPP which only takes a single scale (i.e., only one value of $N$ for the $N \times N$ grid) to partition the proposal into fixed number of divisions, and also backpropagated error signals to the convolution kernels. After feature extraction, feature vectors were fed into a sequence of fully connected layers before two sibling output layers: classification layer(cls) and regression layer(reg). Classification layer was responsible for generating softmax probabilities over C+1 classes(C classes plus one background class), while regression layer encoded 4 real-valued parameters to refine bounding boxes. 
In Fast RCNN, the feature extraction, region classification and bounding box regression steps can all be optimized end-to-end, without extra cache space to store features (unlike SPP Net). Fast R-CNN achieved a much better detection accuracy than R-CNN and SPP-net, and had a better training and inference speed.

Despite the progress in learning detectors, the proposal generation step still relied on traditional methods such as Selective Search~\cite{uijlings2013selective} or Edge Boxes~\cite{zitnick2014edge}, which were based on low-level visual cues and could not be learned in a data-driven manner. To address this issue, \textbf{Faster R-CNN} \cite{ren2015faster} was developed which relied on a novel proposal generator: Region Proposal Network(RPN). This proposal generator could be learned via supervised learning methods. RPN is a fully convolutional network which takes an image of arbitrary size and generates a set of object proposals on each position of the feature map. The network slid over the feature map using an $n\times n$ sliding window, and generated a feature vector for each position. The feature vector was then fed into two sibling output branches, object classification layer (which classified whether the proposal was an object or not) and bounding box regression layer. These results were then fed into the final layer for the actual object classification and bounding box localization. RPN could be inserted into Fast R-CNN and thus the whole framework could be optimized in an end-to-end manner on training data. This way RPN enabled proposal generation in a data driven manner, and was also able to enjoy the discriminative power of deep backbone networks. Faster R-CNN was able to make predictions at 5FPS on GPU and achieved state-of-the-art results on many public benchmark datasets, such as Pascal VOC 2007, 2012 and MSCOCO. Currently, there are huge number of detector variants based on Faster R-CNN for different usage~\cite{cai2018cascade,lin2016fpn,kong2016hypernet,bell2016inside}.

Faster R-CNN computed feature map of the input image and extracted region features on the feature map, which shared feature extraction computation across different regions. However, the computation was not shared in the region classification step, where each feature vector still needed to go through a sequence of FC layers separately. Such extra computation could be extremely large as each image may have hundreds of proposals. Simply removing the fully connected layers would result in the drastic decline of detection performance, as the deep network would have reduced the spatial information of proposals. Dai et al.~\cite{dai2016r} proposed Region-based Fully Convolutional Networks (\textbf{R-FCN}) which shared the computation cost in the region classification step. R-FCN generated a Position Sensitive Score Map which encoded relative position information of different classes, and used a Position Sensitive ROI Pooling layer (PSROI Pooling) to extract spatial-aware region features by encoding each relative position of the target regions. The extracted feature vectors maintained spatial information and thus the detector achieved competitive results compared to Faster R-CNN without region-wise fully connected layer operations.

Another issue with Faster R-CNN was that it used a single deep layer feature map to make the final prediction. This made it difficult to detect objects at different scales. In particular, it was difficult to detect small objects. In DCNN feature representations, deep layer features are semantically-strong but spatially-weak, while shallow layer features are semantically-weak but spatially-strong. Lin et al.~\cite{lin2016fpn} exploited this property and proposed Feature Pyramid Networks(\textbf{FPN}) which combined deep layer features with shallow layer features to enable object detection in feature maps at different scales. The main idea was to strengthen the spatially strong shallow layer features with rich semantic information from the deeper layers. FPN achieved significant progress in detecting multi-scale objects and has been widely used in many other domains such as video detection~\cite{kang2016object,han2016seq} and human pose recognition ~\cite{rayat2018exploiting,pavlakos2017coarse}.

Most instance segmentation algorithms are extended from vanilla object detection algorithms. Early methods~\cite{pinheiro2016learning,pinheiro2015learning,dai2016instance} commonly generated segment proposals, followed by Fast RCNN for segments classification. Later, Dai et al.~\cite{dai2016instance} proposed a multi-stage algorithm named ``MNC" which divided the whole detection framework into multiple stages and predicted segmentation masks from the learned bounding box proposals, which were later categorized by region classifiers. These early works performed bbox and mask prediction in multiple stages. To make the whole process more flexible, He et al.~\cite{he2017mask} proposed \textbf{Mask R-CNN}, which predicted bounding boxes and segmentation masks in parallel based on the proposals and reported state-of-the-art results. Based on Mask R-CNN, Huang et al.~\cite{huang2019mask} proposed a mask-quality aware framework, named Mask Scoring R-CNN, which learned the quality of the predicted masks and calibrated the misalignment between mask quality and mask confidence score. 

\begin{figure*}[thp]
\begin{center}
\includegraphics[width=1.0\linewidth]{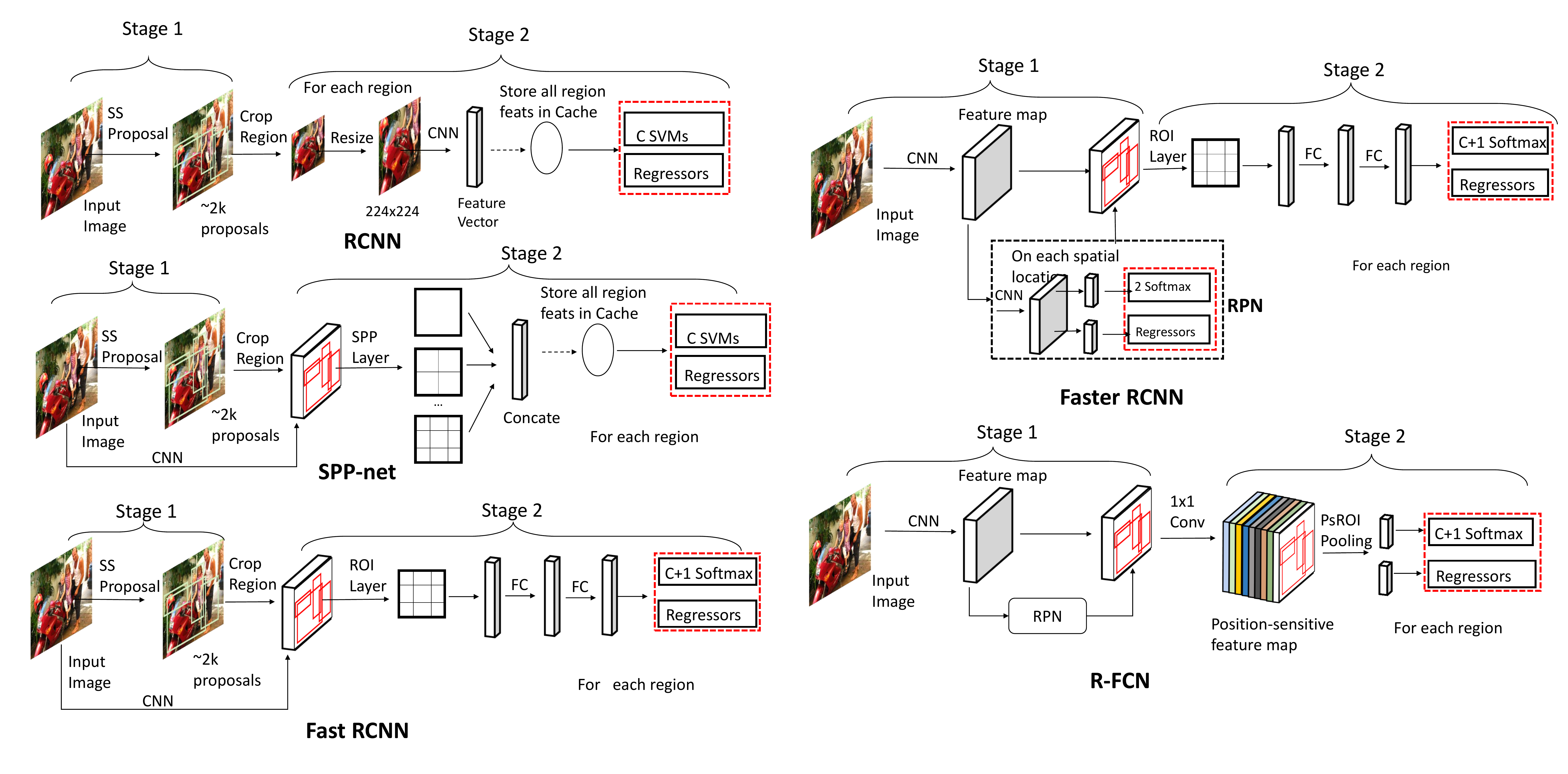}
\end{center}\vspace{-8mm}
\caption{Overview of different two-stage detection frameworks for generic object detection. Red dotted rectangles denote the outputs that define the loss functions. 
}
\label{fig:whole-2}
\end{figure*}

Figure \ref{fig:whole-2} gives an overview of the detection frameworks for several representative two-stage detectors. 

\subsubsection{One-stage Detectors}
Different from two-stage detection algorithms which divide the detection pipeline into two parts: proposal generation and region classification; one-stage detectors do not have a separate stage for proposal generation (or learning a proposal generation). They typically consider all positions on the image as potential objects, and try to classify each region of interest as either background or a target object. 

One of the early successful one-stage detectors based on deep learning was developed by Sermanet et al.~\cite{sermanet2013overfeat} named \textbf{OverFeat}. OverFeat performed object detection by casting DCNN classifier into a fully convolutional object detector. Object detection can be viewed as a "multi-region classification" problem, and thus OverFeat extended the original classifier into detector by viewing the last FC layers as 1x1 convolutional layers to allow arbitrary input. The classification network output a grid of predictions on each region of the input to indicate the presence of an object. After identifying the objects, bounding box regressors were learned to refine the predicted regions based on the same DCNN features of classifier. In order to detect multi-scale objects, the input image was resized into multiple scales which were fed into the network. Finally, the predictions across all the scales were merged together. OverFeat showed significant speed strength compared with RCNN by sharing the computation of overlapping regions using convolutional layers, and only a single pass forward through the network was required. However, the training of classifiers and regressors were separated without being jointly optimized.

Later, Redmon et al.~\cite{redmon2016you} developed a real-time detector called \textbf{YOLO} (You Only Look Once). YOLO considered object detection as a regression problem and spatially divided the whole image into fixed number of grid cells (e.g. using a $7\times7$ grid). Each cell was considered as a proposal to detect the presence of one or more objects. In the original implementation, each cell was considered to contain the center of (upto) two objects. For each cell, a prediction was made which comprised the following information: whether that location had an object, the bounding box coordinates and size(width and height), and the class of the object. The whole framework was a single network and it omitted proposal generation step which could be optimized in an end-to-end manner. Based on a carefully designed lightweight architecture, YOLO could make prediction at 45 FPS, and reach 155 FPS with a more simplified backbone. However, YOLO faced some challenges: 
i) it could detect upto only two objects at a given location, which made it difficult to detect small objects and crowded objects~\cite{redmon2016you}. 
ii) only the last feature map was used for prediction, which was not suitable for predicting objects at multiple scales and aspect ratios.

In 2016, Liu et al. proposed another one-stage detector Single-Shot Mulibox Detector (\textbf{SSD})~\cite{liu2016ssd} which addressed the limitations of YOLO. SSD also divided images into grid cells, but in each grid cell, a set of anchors with multiple scales and aspect-ratios were generated to discretize the output space of bounding boxes (unlike predicting from fixed grid cells adopted in YOLO). Each anchor was refined by 4-value offsets learned by the regressors and was assigned (C+1) categorical probabilities by the classifiers. In addition, SSD predicted objects on multiple feature maps, and each of these feature maps was responsible for detecting a certain scale of objects according to its receptive fields. In order to detect large objects and increase receptive fields, several extra convolutional feature maps were added to the original backbone architecture. The whole network was optimized with a weighted sum of localization loss and classification loss over all prediction maps via an end-to-end training scheme. The final prediction was made by merging all detection results from different feature maps. In order to avoid huge number of negative proposals dominating training gradients, hard negative mining was used to train the detector. Intensive data augmentation was also applied to improve detection accuracy. SSD achieved comparable detection accuracy with Faster R-CNN but enjoyed the ability to do real-time inference.

Without proposal generation to filter easy negative samples, the class imbalance between foreground and background is a severe problem in one-stage detector. Lin et al.~\cite{lin2017focal} proposed a one-stage detector \textbf{RetinaNet} which addressed class imbalance problem in a more flexible manner. RetinaNet used focal loss which suppressed the gradients of easy negative samples instead of simply discarding them. Further, they used feature pyramid networks to detect multi-scale objects at different levels of feature maps. Their proposed focal loss outperformed naive hard negative mining strategy by large margins.

Redmon et al. proposed an improved YOLO version, \textbf{YOLOv2}~\cite{redmon2017yolo9000} which significantly improved detection performance but still maintained real-time inference speed. YOLOv2 adopted a more powerful deep convolutional backbone architecture which was pre-trained on higher resolution images from ImageNet(from $224 \times 224$ to $448 \times 448$), and thus the weights learned were more sensitive to capturing fine-grained information. In addition, inspired by the anchor strategy used in SSD, YOLOv2 defined better anchor priors by k-means clustering from the training data (instead of setting manually). This helped in reducing optimizing difficulties in localization. Finally integrating with Batch Normalization layers~\cite{ioffe2015batch} and multi-scale training techniques, YOLOv2 achieved state-of-the-art detection results at that time. 

The previous approaches required designing anchor boxes manually to train a detector. Later a series of anchor-free object detectors were developed, where the goal was to predict keypoints of the bounding box, instead of trying to fit an object to an anchor. Law and Deng proposed a novel anchor-free framework \textbf{CornerNet}~\cite{law2018cornernet} which detected objects as a pair of corners. On each position of the feature map, class heatmaps, pair embeddings and corner offsets were predicted. Class heatmaps calculated the probabilities of being corners, and corner offsets were used to regress the corner location. And the pair embeddings served to group a pair of corners which belong to the same objects. Without relying on manually designed anchors to match objects, CornerNet obtained significant improvement on MSCOCO datasets. Later there were several other variants of keypoint detection based one-stage detectors~\cite{zhou2019objects,duan2019centernet}.

\begin{figure*}[thp]
\begin{center}
\includegraphics[width=1.\linewidth]{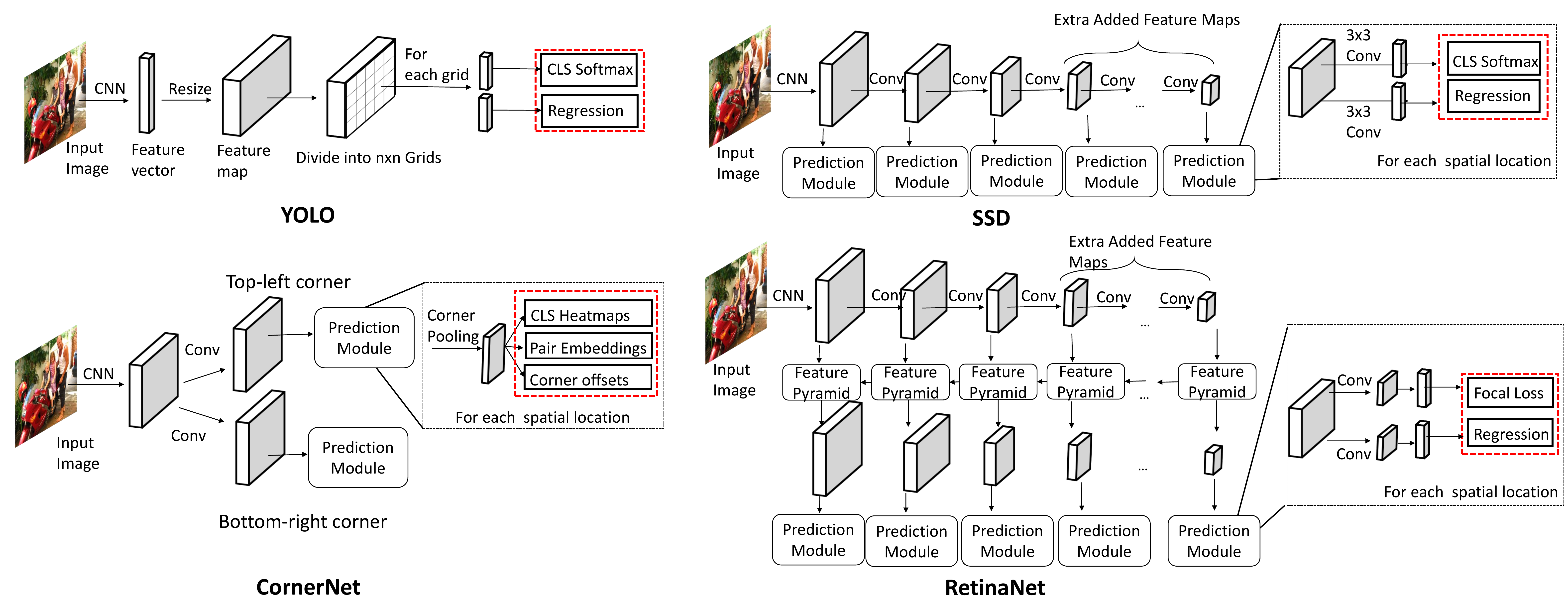}
\end{center}\vspace{-6mm}
\caption{Overview of different one-stage detection frameworks for generic object detection. Red rectangles denotes the outputs that define the objective functions.}
\label{fig:whole-1}
\end{figure*}

Figures \ref{fig:whole-1} gives an overview of different detection frameworks for several representative one-stage detectors. 

\subsection{Backbone Architecture}
R-CNN \cite{girshick2014rich} showed adopting convolutional weights from models pre-trained on large scale image classification problem could provide richer semantic information to train detectors and enhanced the detection performance. During the later years, this approach had become the default strategy for most object detectors. In this section, we will first briefly introduce the basic concept of deep convolutional neural networks and then review some architectures which are widely used for detection.

\subsubsection{Basic Architecture of a CNN}
Deep convolutional neural network (DCNN) is a typical deep neural network and has proven extremely effective in visual understanding~\cite{lecun1998gradient,krizhevsky2012imagenet}. Deep convolutional neural networks are commonly composed of a sequence of convolutional layers, pooling layers, nonlinear activation layers and fully connected layers (FC layers). Convolutional layer takes an image input and convolves over it by $n \times n$ kernels to generate a feature map. The generated feature map can be regarded as a multi-channel image and each channel represents different information about the image. Each pixel in the feature map (named neuron) is connected to a small portion of adjacent neurons from the previous map, which is called the receptive field. After generating feature maps, a non-linear activation layer is applied. Pooling layers are used to summarize the signals within the receptive fields, to enlarge receptive fields as well as reduce computation cost, .

With the combination of a sequence of convolutional layers, pooling layers and non-linear activation layers, the deep convolutional neural network is built. The whole network can be optimized via a defined loss function by gradient-based optimization method (stochastic gradient descent~\cite{robbins1951stochastic}, Adam~\cite{kingma2014adam}, etc.). A typical convolutional neural network is AlexNet~\cite{krizhevsky2012imagenet}, which contains five convolutional layers, three max-pooling layers and three fully connected layers. Each convolutional layer is followed by a ReLU~\cite{nair2010rectified} non-linear activation layer.

\subsubsection{CNN Backbone for Object Detection}
In this section, we will review some architectures which are widely used in object detection tasks with state-of-the-art results, such as VGG16~\cite{ren2015faster,girshick2015fast}, ResNet~\cite{he2016deep,dai2016r}, ResNeXt~\cite{lin2017focal} and Hourglass~\cite{law2018cornernet}. 

VGG16~\cite{simonyan2014very} was developed based on AlexNet. VGG16 is composed of five groups of convolutional layers and three FC layers. There are two convolutional layers in the first two groups and three convolutional layers in the next three groups. Between each group, a Max Pooling layer is applied to decrease spatial dimension. 
VGG16 showed that increasing depth of networks by stacking convolutional layers could increase the model's expression capability, and led to a better performance. However, increasing model depth to 20 layers by simply stacking convolutional layers led to optimization challenges with SGD. The performance declined significantly and was inferior to shallower models, even during the training stages. Based on this observation, He et al.~\cite{he2016deep} proposed ResNet which reduced optimization difficulties by introducing shortcut connections. Here, a layer could skip the nonlinear transformation and directly pass the values to the next layer as is (thus giving us an implicit identity layer). This is given as:
\begin{align}
x_{l+1} =x_{l}+f_{l+1}(x_{l}, \theta)
\end{align}
where $x_{l}$ is the input feature in $l$-th layer and $f_{l+1}$ denotes operations on input $x_{l}$ such as convolution, normalization or non-linear activation. $f_{l+1}(x_{l}, \theta)$ is the residual function to $x_{l}$, so the feature map of any deep layer can be viewed as the sum of the activation of shallow layer and the residual function. Shortcut connection creates a highway which directly propagates the gradients from deep layers to shallow units and thus, significantly reduces training difficulty. 
With residual blocks effectively training networks, the model depth could be increased (e.g. from 16 to 152), allowing us to train very high capacity models. 
Later, He et al.~\cite{he2016identity} proposed a pre-activation variant of ResNet, named ResNet-v2.  
Their experiments showed appropriate ordering of the Batch Normalization ~\cite{ioffe2015batch} could further perform better than original ResNet. This simple but effective modification of ResNet made it possible to successfully train a network with more than 1000 layers, and still enjoyed improved performance due to the increase in depth. Huang et al. argued that although ResNet reduced the training difficulty via shortcut connection, it did not fully utilize features from previous layers. The original features in shallow layers were missing in element-wise operation and thus could not be directly used later. They proposed DenseNet~\cite{huang2017densely}, which retained the shallow layer features, and improved information flow, by concatenating the input with the residual output instead of element-wise addition:
\begin{align}
x_{l+1} =x_{l} \circ f_{l+1}(x_{l}, \theta)
\end{align}
where $\circ$ denotes concatenation. Chen~\cite{chen2017dual} et al. argued that in DenseNet, the majority of new exploited features from shallow layers were duplicated and incurred high computation cost. Integrating the advantages of both ResNet and DenseNet, they propose a Dual Path Network(DPN) which divides $x_l$ channels into two parts: $x_{l}^{d}$ and $x_{l}^{r}$. $x_{l}^{d}$ was used for dense connection computation and $x_{l}^{r}$ was used for element-wise summation, with unshared residual learning branch $f_{l+1}^d$ and $f_{l+1}^r$. The final result was the concatenated output of the two branches:
\begin{align}
x_{l+1} = (x_{l}^{r} + f_{l+1}^r(x_{l}^{r}, \theta^{r})) \circ (x_{l}^{d} \circ f_{l+1}^d(x_{l}^{d}, \theta^d))
\end{align}

Based on ResNet, Xie et al.~\cite{xie2017aggregated} proposed ResNeXt which considerably reduced computation and memory cost while maintaining comparable classification accuracy. 
ResNeXt adopted group convolution layers~\cite{krizhevsky2012imagenet} which sparsely connects feature map channels to reduce computation cost. By increasing group number to keep computation cost consistent to the original ResNet, ResNeXt captures richer semantic feature representation from the training data and thus improves backbone accuracy.
Later, Howard et al.~\cite{howard2017mobilenets} set the coordinates equal to number of channels of each feature map and developed MobileNet. MobileNet significantly reduced computation cost as well as number of parameters without significant loss in classification accuracy. This model was specifically designed for usage on a mobile platform. 

In addition to increasing model depth, some efforts explored benefits from increasing model width to improve the learning capacity. Szegedy et al. proposed GoogleNet with an inception module~\cite{szegedy2015going} which applied different scale convolution kernels ($1 \times 1,3\times 3$ and $5 \times 5$) on the same feature map in a given layer. This way it captured multi-scale features and summarized these features together as an output feature map. Better versions of this model were developed later with different design of choice of convolution kernels ~\cite{szegedy2016rethinking}, and introducing residual blocks~\cite{szegedy2017inception}. 

The network structures introduced above were all designed for image classification. Typically these models trained on ImageNet are adopted as initialization of the model used for object detection. However, directly applying this pre-trained model from classification to detection is sub-optimal due to a potential conflict between classification and detection tasks. Specifically, i) classification requires large receptive fields and wants to maintain spatial invariance. Thus multiple downsampling operation (such as pooling layer) are applied to decrease feature map resolution. The feature maps generated are low-resolution and spatially invariant and have large receptive fields. However, in detection, high-resolution spatial information is required to correctly localize objects; and
ii) classification makes predictions on a single feature map, while detection requires feature maps with multiple representations to detect objects at multiple scales. 
To bridge the difficulties between the two tasks, Li et al. introduced DetNet~\cite{li2018detnet} which was designed specifically for detection. DetNet kept high resolution feature maps for prediction with dilated convolutions to increase receptive fields. In addition, DetNet detected objects on multi-scale feature maps, which provided richer information. DetNet was pre-trained on large scale classification dataset while the network structure was designed for detection.

Hourglass Network~\cite{newell2016stacked} is another architecture, which was not designed specifically for image classification. Hourglass Network first appeared in human pose recognition task~\cite{newell2016stacked}, and was a fully convolutional structure with a sequence of hourglass modules. Hourglass module first downsampled the input image via a sequence of convolutional layer or pooling layer, and upsampled the feature map via deconvolutional operation. To avoid information loss in downsampling stage, skip connection were used between downsampling and upsampling features. Hourglass module could capture both local and global information and thus was very suitable for object detection. Currently Hourglass Network is widely used in state-of-the-art detection frameworks~\cite{law2018cornernet,zhou2019objects,duan2019centernet}.

\subsection{Proposal Generation}\label{sec:prop}
Proposal generation plays a very important role in the object detection framework. A proposal generator generates a set of rectangle bounding boxes, which are potentially objects. These proposals are then used for classification and localization refinement. We categorize proposal generation methods into four categories: traditional computer vision methods, anchor-based supervised learning methods, keypoint based methods and other methods. Notably, both one-stage detectors and two-stage detectors generate proposals, the main difference is two-stage detectors generates a sparse set of proposals with only foreground or background information, while one-stage detectors consider each region in the image as a potential proposal, and accordingly estimates the class and bounding box coordinates of potential objects at each location. 

\subsubsection{Traditional Computer Vision Methods}
These methods generate proposals in images using traditional computer vision methods based on low-level cues, such as edges, corners, color, etc. These techniques can be categorized into three principles: i) computing the 'objectness score' of a candidate box; ii) merging super-pixels from original images; iii) generating multiple foreground and background segments;

\textbf{Objectness Score} based methods predict an objectness score of each candidate box measuring how likely it may contain an object. Arbelaez et al.~\cite{alexe2012measuring} assigned objectness score to proposals by classification based on visual cues such as color contrast, edge density and saliency. Rahtu et al.\cite{rahtu2011learning} revisited the idea of Arbelaez et al.~\cite{alexe2012measuring} and introduced a more efficient cascaded learning method to rank the objectness score of candidate proposals. 

\textbf{Superpixels Merging} is based on merging superpixels generated from segmentation results. 
Selective Search~\cite{uijlings2013selective} was a proposal generation algorithm based on merging super-pixels. It computed the multiple hierarchical segments generated by segmentation method~\cite{felzenszwalb2004efficient}, which were merged according to their visual factors(color, areas, etc.), and finally bounding boxes were placed on the merged segments. Manen et al.~\cite{manen2013prime} proposed a similar idea to merge superpixels. The difference was that the weight of the merging function was learned and the merging process was randomized. Selective Search is widely used in many detection frameworks due to its efficiency and high recall compared to other traditional methods. 

\textbf{Seed Segmentation} starts with multiple seed regions, and for each seed, foreground and background segments are generated. To avoid building up hierarchical segmentation, CPMC~\cite{carreira2011cpmc} generated a set of overlapping segments initialized with diverse seeds. Each proposal segment was the solution of a binary(foreground or background) segmentation problem. Enreds and Hoiem~\cite{endres2014category} combined the idea of Selective Search~\cite{uijlings2013selective} and CPMC~\cite{carreira2011cpmc}. It started with super-pixels and merged them with new designed features. These merged segments were used as seeds to generate larger segments, which was similar to CPMC. However, producing high quality segmentation masks is very time-consuming and it's not applicable to large scale datasets. 

The primary advantage of these traditional computer vision methods is that they are very simple and can generate proposals with high recall (e.g. on medium scale datasets such as Pascal VOC). 
However, these methods are mainly based on low level visual cues such as color or edges. They cannot be jointly optimized with the whole detection pipeline. Thus they are unable to exploit the power of large scale datasets to improve representation learning. 
On challenging datasets such as MSCOCO~\cite{lin2014microsoft}, traditional computer vision methods struggled to generate high quality proposals due to these limitations.

\subsubsection{Anchor-based Methods}
\begin{figure}[hpt]
\begin{center}
\includegraphics[width=0.8\linewidth]{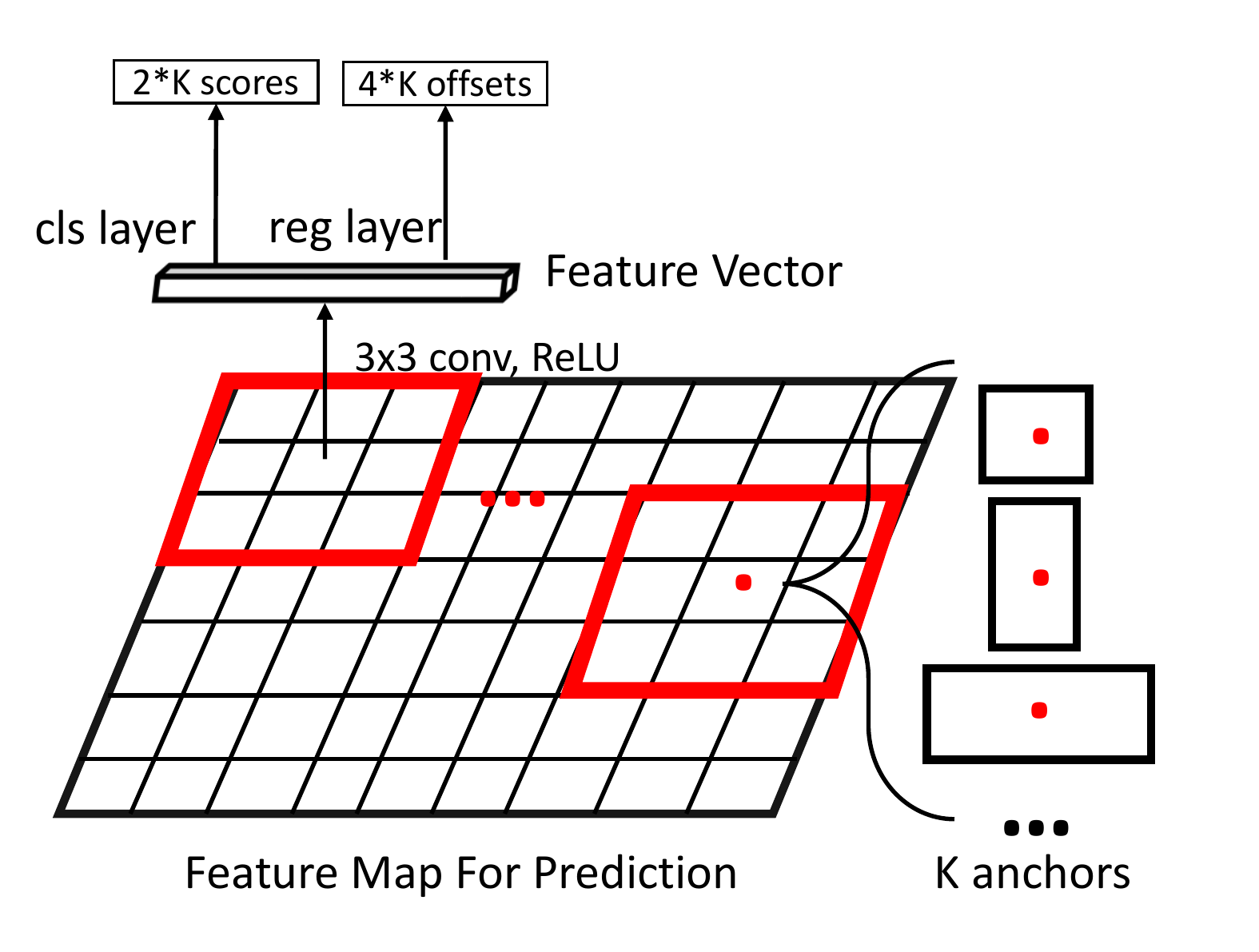}
\end{center}
\caption{Diagram of RPN~\cite{ren2015faster}. Each position of the feature map connects with a sliding windows, followed with two sibling branches.}
\label{fig:rpn}
\end{figure}

One large family of supervised proposal generators is anchor-based methods. They generate proposals based on pre-defined anchors. Ren et al. proposed Region Proposal Network (RPN)~\cite{ren2015faster} to generate proposals in a supervised way based on deep convolutional feature maps. The network slid over the entire feature map using $3 \times 3$ convolution filters.
For each position, $k$ anchors (or initial estimates of bounding boxes) of varying size and aspect ratios were considered. These sizes and ratios allowed for matching objects at different scales in the entire image. 
Based on the ground truth bonding boxes, the object locations were matched with the most appropriate anchors to obtain the supervision signal for the anchor estimation. 
A $256-$dimensional feature vector was extracted from each anchor and was fed into two sibling branches - classification layer and regression layer. Classification branch was responsible for modeling objectness score while regression branch encoded four real-values to refine location of the bounding box from the original anchor estimation. Based on the ground truth, each anchor was predicted to either be an object, or just background by the classification branch (See Fig. \ref{fig:rpn}).
Later, SSD~\cite{liu2016ssd} adopted a similar idea of anchors in RPN by using multi-scale anchors to match objects. 
The main difference was that SSD assigned categorical probabilities to each anchor proposal, while RPN first evaluated whether the anchor proposal was foreground or background and performed the categorical classification in the next stage.

Despite promising performance, the anchor priors are manually designed with multiple scales and aspect ratios in a heuristic manner. These design choices may not be optimal, and different datasets would require different anchor design strategies. Many efforts have been made to improve the design choice of anchors. Zhang et al. proposed Single Shot Scale-invariant Face Detector (S3FD)~\cite{Zhang_2017_ICCV} based on SSD with carefully designed anchors to match the objects. According to the effective receptive field~\cite{luo2016understanding} of different feature maps, different anchor priors were designed. Zhu et al.~\cite{zhu2018seeing} introduced an anchor design method for matching small objects by enlarging input image size and reducing anchor strides. Xie et al. proposed Dimension-Decomposition Region Proposal Network (DeRPN)~\cite{xie2019DeRPN} which decomposed the dimension of anchor boxes based on RPN. DeRPN used an anchor string mechanism to independently match objects width and height. This helped match objects with large scale variance and reduced the searching space.

Ghodrati et al. developed DeepProposals~\cite{ghodrati2015deepproposal} which predicted proposals on the low-resolution deeper layer feature map. These were then projected back onto the high-resolution shallow layer feature maps, where they are further refined. Redmon et al.~\cite{redmon2017yolo9000} designed anchor priors by learning priors from the training data using k-means clustering. Later, Zhang et al. introduced Single-Shot Refinement Neural Network (RefineDet)~\cite{zhang2018single} which refined the manually defined anchors in two steps. In the first step, RefineDet learned a set of localization offsets based on the original hand-designed anchors and these anchors were refined by the learned offsets. In the second stage, a new set of localization offsets were learned based on the refined anchors from the first step for further refinement. This cascaded optimization framework significantly improved the anchor quality and final prediction accuracy in a data-driven manner. Cai et al. proposed Cascade R-CNN~\cite{cai2018cascade} which adopted a similar idea as RefineDet by refining proposals in a cascaded way. Yang et al.~\cite{NIPS2018_7315} modeled anchors as functions implemented by neural networks which was computed from customized anchors. Their method MetaAnchor showed comprehensive improvement compared to other manually defined methods but the customized anchors were still designed manually. 

\subsubsection{Keypoints-based Methods}
Another proposal generation approach is based on keypoint detection, which can be divided into two families: corner-based methods and center-based methods. \textit{{Corner-based methods}} predict bounding boxes by merging pairs of corners learned from the feature map. Denet~\cite{tychsen2017denet} reformulated the object detection problem in a probabilistic way. For each point on the feature map, Denet modeled the distribution of being one of the 4 corner types of objects (top-left, top-right, bottom-left, bottom-right), and applied a naive bayesian classifiers over each corner of the objects to estimate the confidence score of a bounding box. This corner-based algorithm eliminated the design of anchors and became a more effective method to produce high quality proposals. Later based on Denet, Law and Deng proposed CornerNet~\cite{law2018cornernet} which directly modeled categorical information on corners. CornerNet modeled information of top-left and bottom-right corners with novel feature embedding methods and corner pooling layer to correctly match keypoints belonging to the same objects, obtaining state-of-the-art results on public benchmarks. For \textit{center-based methods}, the probability of being the center of the objects is predicted on each position of the feature map, and the height and width are directly regressed without any anchor priors. Zhu et al.~\cite{zhu2019feature} presented a feature-selection-anchor-free (FSAF) framework which could be plugged into one-stage detectors with FPN structure. In FSAF, an online feature selection block is applied to train multi-level center-based branches attached in each level of the feature pyramid. During training, FSAF dynamically assigned each object to the most suitable feature level to train the center-based branch. Similar to FSAF, Zhou et al. proposed a new center-based framework~\cite{zhou2019objects} based on a single Hourglass network~\cite{law2018cornernet} without FPN structure. Furthermore, they applied center-based method into higher-level problems such as 3D-detection and human pose recognition, and all achieved state-of-the-art results. Duan et al.~\cite{duan2019centernet} proposed CenterNet, which combined the idea of center-based methods and corner-based methods. CenterNet first predicted bounding boxes by pairs of corners, and then predicted center probabilities of the initial prediction to reject easy negatives. CenterNet obtained significant improvements compared with baselines. These anchor-free methods form a promising research direction in the future.

\subsubsection{Other Methods}
There are some other proposal generation algorithms which are not based on keypoints or anchors but also offer competitive performances. Lu et al. proposed AZnet~\cite{lu2016adaptive} which automatically focused on regions of high interest. AZnet adopted a search strategy that adaptively directed computation resources to sub-regions which were likely contain objects. For each region, AZnet predicted two values: zoom indicator and adjacency scores. Zoom indicator determined whether to further divide this region which may contain smaller objects and adjacency scores denoted its objectness. The starting point was the entire image and each divided sub-region is recursively processed in this way until the zoom indicator is too small. AZnet was better at matching sparse and small objects compared to RPN's anchor-object matching approach.

\subsection{Feature Representation Learning}
Feature Representation Learning is a critical component in the whole detection framework. Target objects lie in complex environments and have large variance in scale and aspect ratios. There is a need to train a robust and discriminative feature embedding of objects to obtain a good detection performance. In this section, we introduce feature representation learning strategies for object detection. Specifically, we identify three categories: multi-scale feature learning, contextual reasoning, and deformable feature learning.

\subsubsection{Multi-scale Feature Learning}
\begin{figure}[htb]
\centering
\includegraphics[width=3.4in]{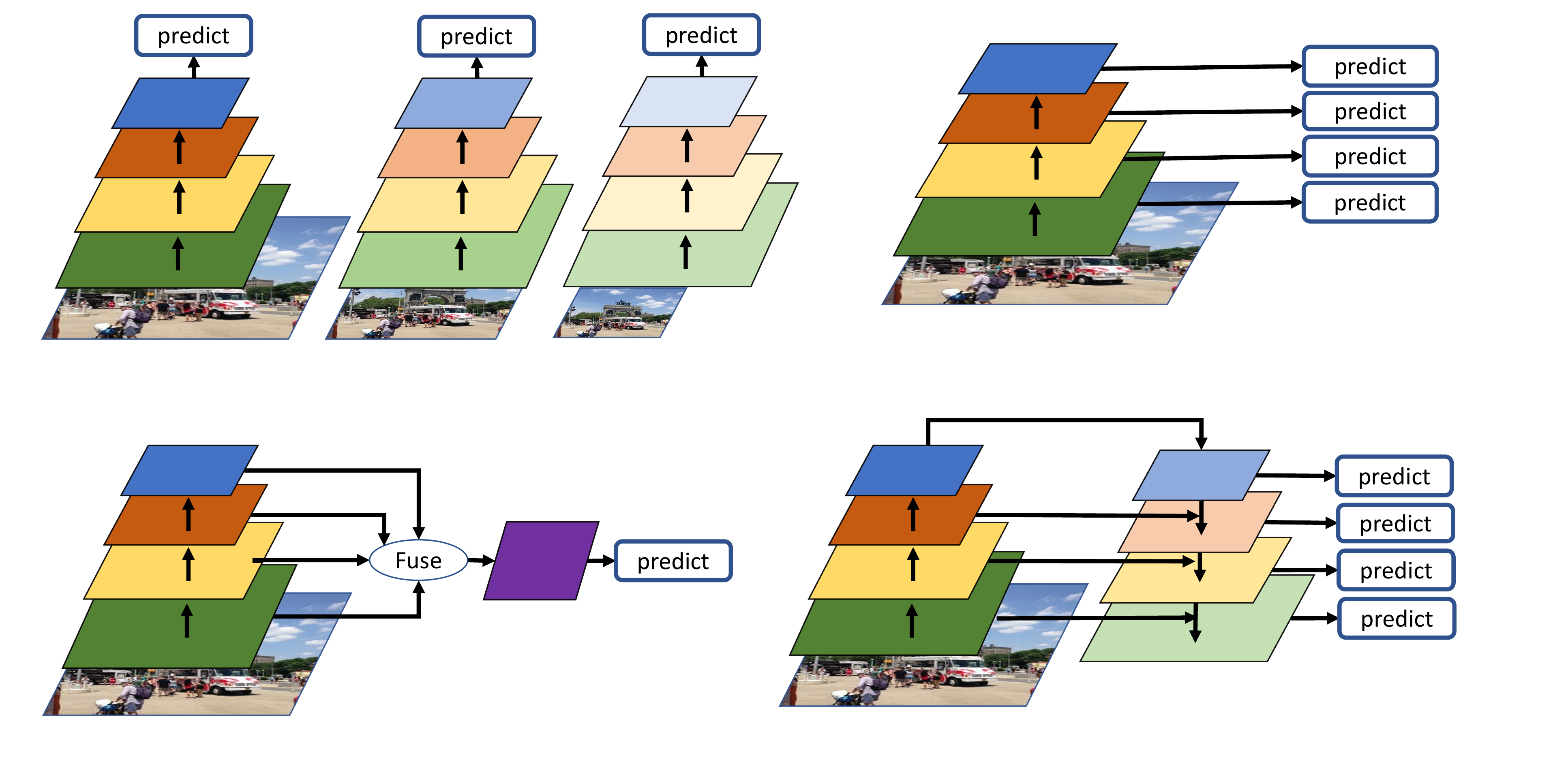}
\caption{Four paradigms for multi-scale feature learning. Top Left: \textbf{Image Pyramid}, which learns multiple detectors from different scale images; Top Right: \textbf{Prediction Pyramid}, which predicts on multiple feature maps;
Bottom Left: \textbf{Integrated Features}, which predicts on single feature map generated from multiple features; Bottom Right: \textbf{Feature Pyramid} which combines the structure of \textbf{Prediction Pyramid} and \textbf{Integrated Features}.}
\label{fig:scale}
\end{figure}

Typical object detection algorithms based on deep convolutional networks such as Fast R-CNN~\cite{girshick2015fast} and Faster R-CNN~\cite{ren2015faster} use only a single layer's feature map to detect objects. However, detecting objects across large range of scales and aspect ratios is quite challenging on a single feature map. Deep convolutional networks learn hierarchical features in different layers which capture different scale information. Specifically, shallow layer features with spatial-rich information have higher resolution and smaller receptive fields and thus are more suitable for detecting small objects, while semantic-rich features in deep layers are more robust to illumination, translation and have larger receptive fields (but coarse resolutions), and are more suitable for detecting large objects. When detecting small objects, high resolution representations are required and the representation of these objects may not even be available in the deep layer features, making small object detection difficult. 
Some techniques such as dilated/atrous convolutions~\cite{dai2017deformable,dai2016r} were proposed to avoid downsampling, and used the high resolution information even in the deeper layers.
At the same time, detecting large objects in shallow layers are also non-optimal without large enough receptive fields. Thus, handling feature scale issues has become a fundamental research problem within object detection. There are four main paradigms addressing multi-scale feature learning problem:
Image Pyramid, Prediction Pyramid, Integrated Features and Feature Pyramid. These are briefly illustrated in the Fig. \ref{fig:scale}.

\begin{figure}[htb]
\centering
\includegraphics[width=2.0in]{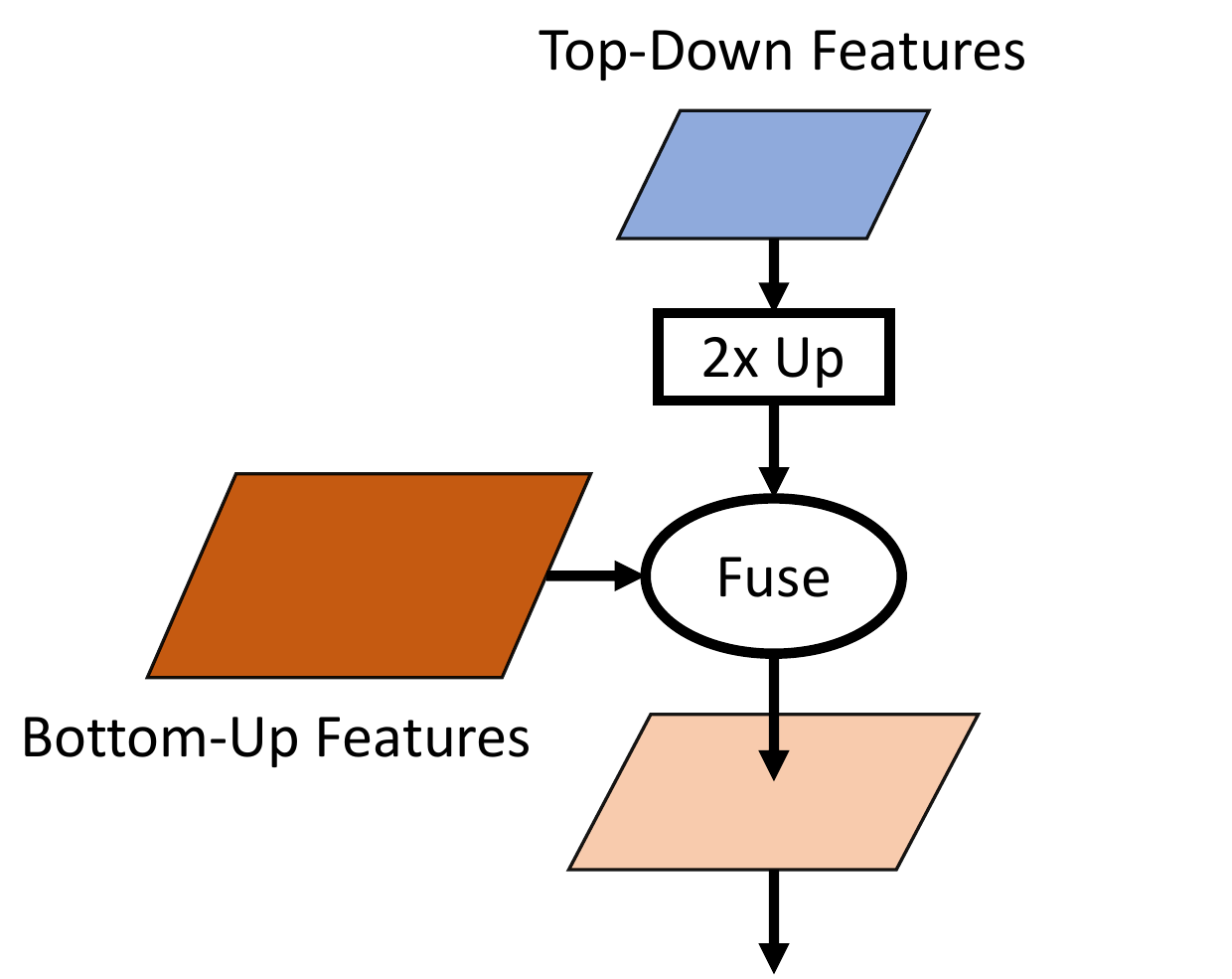}
\caption{General framework for feature combination. Top-down features are 2 times up-sampled and fuse with bottom-up features. The fuse methods can be element-wise
sum, multiplication, concatenation and so on. Convolution and normalization layers can be inserted in to this general framework to enhance semantic information and reduce
memory cost.}
\label{fig:fpn}
\end{figure}

\textbf{Image Pyramid}: An intuitive idea is to resize input images into a number of different scales (Image Pyramid) and to train multiple detectors, each of which is responsible for a certain range of scales~\cite{singh2018analysis,Hu_2017_CVPR,yang2016exploit,liu_2017_rsa}. During testing, images are resized to different scales followed by multiple detectors and the detection results are merged. This can be computationally expensive. Liu et al.~\cite{liu_2017_rsa} first learned a light-weight scale-aware network to resize images such that all objects were in a similar scale. This was followed by learning a single scale detector. Singh et. al.~\cite{singh2018analysis} conducted comprehensive experiments on small object detection. They argued that learning a single scale-robust detector to handle all scale objects was much more difficult than learning scale-dependent detectors with image pyramids. In their work, they proposed a novel framework Scale Normalization for Image Pyramids (SNIP)~\cite{singh2018analysis} which trained multiple scale-dependent detectors and each of them was responsible for a certain scale objects.

\textbf{Integrated Features}: Another approach is to construct a single feature map by combining features in multiple layers and making final predictions based on the new constructed map~\cite{shrivastava2016beyond,bell2016inside,kong2016hypernet,wang2018multi,kim2016pvanet,ronneberger2015u}. By fusing spatially rich shallow layer features and semantic-rich deep layer features, the new constructed features contain rich information and thus can detect objects at different scales. 
These combinations are commonly achieved by using skip connections \cite{he2016deep}. 
Feature normalization is required as feature norms of different layers have a high variance. 
Bell et al. proposed Inside-Outside Network (ION)~\cite{bell2016inside} which cropped region features from different layers via ROI Pooling \cite{girshick2015fast}, and combined these multi-scale region features for the final prediction. Kong et. al. proposed HyperNet~\cite{kong2016hypernet} which adopted a similar idea as IoN. They carefully designed high resolution hyper feature maps by integrating intermediate and shallow layer features to generate proposals and detect objects. Deconvolutional layers were used to up-sample deep layer feature maps and batch normalization layers were used to normalize input blobs in their work. The constructed hyper feature maps could also implicitly encode contextual information from different layers. Inspired by fine-grained classification algorithms which integrate high-order representation instead of exploiting simple first-order representations of object proposals, Wang et al. proposed a novel framework Multi-scale Location-aware Kernel Representation (MLKP)~\cite{wang2018multi} which captured high-order statistics of proposal features and generated more discriminative feature representations efficiently. The combined feature representation was more descriptive and provides both semantic and spatial information for both classification and localization. 

\textbf{Prediction Pyramid}: Liu et al.'s SSD ~\cite{liu2016ssd} combined coarse and fine features from multiple layers together. In SSD, predictions were made from multiple layers, where each layer was responsible for a certain scale of objects. Later, many efforts~\cite{cai2016unified,shen2017dsod,liu2018receptive} followed this principle to detect multi-scale objects. Yang et al.~\cite{yang2016exploit} also exploited appropriate feature maps to generate certain scale of object proposals and these feature maps were fed into multiple scale-dependent classifiers to predict objects. In their work, cascaded rejection classifiers were learned to reject easy background proposals in early stages to accelerate detection speed. Multi-scale Deep Convolutional Neural Network (MSCNN)~\cite{cai2016unified} applied deconvolutional layers on multiple feature maps to improve their resolutions, and later these refined feature maps were used to make predictions. Liu et al. proposed a Receptive Field Block Net (RFBNet)~\cite{liu2018receptive} to enhance the robustness and receptive fields via a receptive field block (RFB block). RFB block adopted similar ideas as the inception module~\cite{szegedy2015going} which captured features from multiple scale and receptive fields via multiple branches with different convolution kernels and finally merged them together.

\textbf{Feature Pyramid}: To combine the advantage of Integrated Features and Prediction Pyramid, Lin et al. proposed Feature Pyramid Network(FPN)~\cite{lin2016fpn} which integrated different scale features with lateral connections in a top-down fashion to build a set of scale invariant feature maps, and multiple scale-dependent classifiers were learned on these feature pyramids. Specifically, the deep semantic-rich features were used to strengthen the shallow spatially-rich features. These top-down and lateral features were combined by element-wise summation or concatenation, with small convolutions reducing the dimensions. FPN showed significant improvement in object detection, as well as other applications, and achieved state-of-the art results in learning multi-scale features. Many variants of FPN were later developed ~\cite{ren2017accurate,jeong2017enhancement,ren2017accurate,zhou2018scale,fu2017dssd,zhang2018single,woo2018stairnet,li2017zoom,kong2018deep,zhao2019m2det,li2017fssd,lee2017residual,cui2018mdssd}, with modifications to the feature pyramid block (see Fig. \ref{fig:fpn}). Kong et al.~\cite{kong2017ron} and Zhang et. al.~\cite{zhang2018single} built scale invariant feature maps with lateral connections. Different from FPN which generated region proposals followed by categorical classifiers, their methods omitted proposal generation and thus were more efficient than original FPN. Ren et al.~\cite{ren2017accurate} and Jeong et al. ~\cite{jeong2017enhancement} developed a novel structure which gradually and selectively encoded contextual information between different layer features. Inspired by super resolution tasks~\cite{lim2017enhanced,shi2016real}, Zhou et al.~\cite{zhou2018scale} developed high resolution feature maps using a novel transform block which explicitly explored the inter-scale consistency nature across multiple detection scales.

\subsubsection{Region Feature Encoding}
For two-stage detectors, region feature encoding is a critical step to extract features from proposals into fixed length feature vectors. In R-CNN, Girshick et al.~\cite{girshick2014rich} cropped region proposals from the whole image and resized the cropped regions into fixed sized patches($224 \times 224$) via bilinear interpolation, followed by a deep convolution feature extractor. Their method encoded high resolution region features but the computation was expensive. 

Later Girshick et al.~\cite{girshick2015fast} and Ren~\cite{ren2015faster} proposed ROI Pooling layer to encode region features. ROI Pooling divided each region into $n \times n$ cells (e.g. $7 \times 7$ by default) and only the neuron with the maximum signal would go ahead in the feedforward stage. This is similar to max-pooling, but across (potentially) different sized regions. ROI Pooling extracted features from the down-sampled feature map and as a result struggled to handle small objects. Dai~\cite{dai2016instance} proposed ROI Warping layer which encoded region features via bilinear interpolation. Due to the downsampling operation in DCNN, there can be a misalignment of the object position in the original image and the downsampled feature maps, which RoI Pooling and RoI Warping layers are not able to handle. Instead of quantizing grids border as ROI Warping and ROI Pooling do, He et al.~\cite{he2017mask} proposed ROI Align layer which addressed the quantization issue by bilinear interpolation at fractionally sampled positions within each grid. Based on ROI Align, Jiang et al.~\cite{jiang2018acquisition} presented Precise ROI Pooing (PrROI Pooling), which avoided any quantization of coordinates and had a continuous gradient on bounding box coordinates. 

In order to enhance spatial information of the downsampled region features, Dai et al.~\cite{dai2016r} proposed Position Sensitive ROI Pooing (PSROI Pooling) which kept relative spatial information of downsampled features. Each channel of generated region feature map only corresponded to a subset channels of input region according to its relative spatial position.
Based on PSROI Pooling, Zhai et al.~\cite{zhai2018feature} presented feature selective networks to learn robust region features by exploiting disparities among sub-region and aspect ratios. The proposed network encoded sub-region and aspect ratio information which were selectively pooled to refine initial region features by a light-weight head.

Later, more algorithms were proposed to well encode region features from different viewpoints. 
Zhu et al. proposed CoupleNet~\cite{zhu2017couplenet} which extracted region features by combining outputs generated from both ROI Pooling layer and PSROI Pooling layer. ROI Pooling layer extracted global region information but struggled for objects with high occlusion while PSROI Pooling layer focused more on local information. CoupleNet enhanced features generated from ROI Pooling and PSROI Pooling by element-wise summation and generated more powerful features. Later Dai et al. proposed Deformable ROI Pooling~\cite{dai2017deformable} which generalized aligned RoI pooling by learning an offset for each grid and adding it to the grid center. The sub-grid start with a regular ROI Pooling layer to extract initial region features and the extracted features were used to regress offset by an auxiliary network. Deformable ROI Pooling can automatically model the image content without being constrained by fixed receptive fields. 

\vspace{-5mm}
\subsubsection{Contextual Reasoning}
Contextual information plays an important role in object detection. Objects often tend to appear in specific environments and sometimes also coexist with other objects. For each example, birds commonly fly in the sky. Effectively using contextual information can help improve detection performance, especially for detecting objects with insufficient cues(small object, occlusion etc.) Learning the relationship between objects with their surrounding context can improve detector's ability to understand the scenario. For traditional object detection algorithms, there have been several efforts exploring context~\cite{galleguillos2010context}, but for object detection based on deep learning, context has not been extensively explored. This is because convolutional networks implicitly already capture contextual information from hierarchical feature representations. However, some recent efforts~ \cite{he2016deep,ouyang2015deepid,chu2018deep,zhu2015segdeepm,he2017mask,dai2016instance,he2017mask,chen2017spatial,gidaris2015object,cai2016unified} still try to exploit contextual information. Some works~\cite{cheng2018revisiting} have even shown that in some cases context information may even harm the detection performance. In this section we review contextual reasoning for object detection from two aspects: \textit{global context} and \textit{region context}. 

\textit{Global context reasoning} refers to learning from the context in the whole image. Unlike traditional detectors which attempt to classify specific regions in the image as objects, the idea here is to use the contextual information (i.e., information from the rest of the image) to classify a particular region of interest. For example, detecting a baseball ball from an image can be challenging for a traditional detector (as it may be confused with balls from other sports); but if the contextual information from the rest of the image is used (e.g. baseball field, players, bat), it becomes easier to identify the baseball ball object. 

Some representative efforts include ION~\cite{bell2016inside}, DeepId~\cite{ouyang2015deepid} and improved version of Faster R-CNN ~\cite{he2016deep}. In ION, Bell et al. used recurrent neural network to encode contextual information across the whole image from four directions. Ouyang et al.~\cite{ouyang2015deepid} learned a categorical score for each image which is used as contextual features concatenated with the object detection results. He et al.~\cite{he2016deep} extracted feature embedding of the entire image and concatenate it with region features to improve detection results. In addition, some methods~\cite{zhu2015segdeepm,he2017mask,dai2016instance,zhao2018pseudo,zhang2018single_2,shrivastava2016contextual,li2018auto} exploit global contextual information via semantic segmentation. Due to precise pixel-level annotation, segmentation feature maps capture strong spatial information. 
He et al.~\cite{he2017mask} and Dai et al.~\cite{dai2016instance} learn unified instance segmentation framework and optimize the detector with pixel-level supervision. They jointly optimized detection and segmentation objectives as a multi-task optimization. Though segmentation can significantly improve detection performance, obtaining the pixel-level annotation is very expensive. Zhao et al.~\cite{zhao2018pseudo} optimized detectors with pseudo segmentation annotation and showed promising results. Zhang et al.'s work Detection with Enriched Semantics (DES)~\cite{zhang2018single_2}, introduced contextual information by learning a segmentation mask without segemtation annotations. It also jointly optimized object detection and segmentation objectives and enriched original feature map with a more discriminative feature map.

\textit{Region Context Reasoning} encodes contextual information surrounding regions and learns interactions between the objects with their surrounding area. Directly modeling different locations and categories objects relations with the contextual is very challenging. Chen et al. proposed Spatial Memory Network (SMN)~\cite{chen2017spatial} which introduced a spatial memory based module. The spatial memory module captured instance-level contexts by assembling object instances back into a pseudo "image" representations which were later used for object relations reasoning. 
Liu et al. proposed Structure Inference Net (SIN)~\cite{liu2018structure} which formulated object detection as a graph inference problem by considering scene contextual information and object relationships. In SIN, each object was treated as a graph node and the relationship between different objects were regarded as graph edges. 
Hu et al.~\cite{hu2018relation} proposed a lightweight framework relation network which formulated the interaction between different objects between their appearance and image locations. The new proposed framework did not need additional annotation and showed improvements in object detection performance. Based on Hu et al., Gu et al.~\cite{gu2018learning} proposed a fully learnable object detector which proposed a general viewpoint that unified existing region feature extraction methods.
Their proposed method removed heuristic choices in ROI pooling methods and automatically select the most significant parts, including contexts beyond proposals. Another method to encode contextual information is to implicitly encode region features by adding image features surrounding region proposals and a large number of approaches have been proposed based on this idea~\cite{gidaris2015object,cai2016unified,xu2018deep,chen2018context,zeng2016gated,li2017attentive}. In addition to encode features from region proposals, Gidaris et al.~\cite{gidaris2015object} extracted features from a number of different sub-regions of the original object proposals(border regions, central regions, contextual regions etc.) and concatenated these features with the original region features. Similar to their method, \cite{cai2016unified} extracted local contexts by enlarging the proposal window size and concatenating these features with the original ones. Zeng et al.~\cite{zeng2016gated} proposed Gated Bi-Directional CNN (GBDNet) which extracted features from multi-scale subregions. Notably, GBDNet learned a gated function to control the transmission of different region information because not all contextual information is helpful for detection.

\subsubsection{Deformable Feature Learning}
A good detector should be robust to nonrigid deformation of objects. Before the deep learning era, Deformable Part based Models(DPMs)~ \cite{felzenszwalb2008discriminatively} had been successfully used for object detection. DPMs represented objects by multiple component parts using a deformable coding method, making the detector robust to nonrigid object transformation. In order to enable detectors based on deep learning to model deformations of object parts, many researchers have developed detection frameworks to explicitly model object parts~\cite{dai2017deformable,ouyang2015deepid,zhu2019deformv2,girshick2015deformable}. DeepIDNet~\cite{ouyang2015deepid} developed a deformable-aware pooling layer to encode the deformation information across different object categories. Dai et al.~\cite{dai2017deformable} and Zhu et al.~\cite{zhu2019deformv2} designed deformable convolutional layers which automatically learned the auxiliary position offsets to augment information sampled in regular sampling locations of the feature map.

\section{Learning Strategy}\label{sec:lrstr}
In contrast to image classification, object detection requires optimizing both localization and classification tasks, which makes it more difficult to train robust detectors. In addition, there are several issues that need to be addressed, such as imbalance sampling, localization, acceleration etc. Thus there is a need to develop innovative learning strategies to train effective and efficient detectors. In this section, we review some of the learning strategies for object detection.

\subsection{Training Stage}
In this section, we review the learning strategies for training object detectors. Specifically we discuss, data augmentation, imbalance sampling, cascade learning, localization refinement and some other learning strategies.

\subsubsection{Data Augmentation.} Data augmentation is important for nearly all deep learning methods as they are often data-hungry and more training data leads to better results. In object detection, in order to increase training data as well as generate training patches with multiple visual properties, Horizontal flips of training images is used in training Faster R-CNN detector~\cite{girshick2015fast}. A more intensive data augmentation strategy is used in one-stage detectors including rotation, random crops, expanding and color jittering~\cite{liu2016ssd,cai2016unified,singh2018sniper}. This data augmentation strategy has shown significant improvement in detection accuracy. 

\subsubsection{Imbalance Sampling}
In object detection, imbalance of negative and positive samples is a critical issue. That is, most of the regions of interest estimated as proposals are in fact just background images. Very few of them are positive instances (or objects). This results in problem of imbalance while training detectors. Specifically, two issues arise, which need to be addressed: class imbalance and difficulty imbalance. The class imbalance issue is that most candidate proposals belong to the background and only a few of proposals contain objects. This results in the background proposals dominating the gradients during training. The difficulty imbalance is closely related to the first issue, where due to the class imbalance, it becomes much easier to classify most of the background proposals easily, while the objects become harder to classify. 
A variety of strategies have been developed to tackle the class imbalance issue. Two-stage detectors such as R-CNN and Fast R-CNN will first reject majority of negative samples and keep 2,000 proposals for further classification. In Fast R-CNN~\cite{girshick2015fast}, negative samples were randomly sampled from these 2k proposals and the ratio of positive and negative was fixed as 1:3 in each mini-batch, to further reduce the adverse effects of class imbalance.
Random sample can address class imbalance issue but are not able to fully utilize information from negative proposals. Some negative proposals may contain rich context information about the images, and some hard proposals can help to improve detection accuracy. To address this, Liu et al.~\cite{liu2016ssd} proposed hard negative sampling strategy which fixed the foreground and background ratio but sampled most difficult negative proposals for updating the model. Specifically, negative proposals with higher classification loss were selected for training.

To address difficulty imbalance, most sampling strategies are based on carefully designed loss functions. For obejct detection, a \textit{multi-class} classifier is learned over C+1 categories(C target categories plus one background category). Assume the region is labeled with ground truth class $u$, and $p$ is the output discrete probability distribution over C+1 classes($p = \left \{ p_0, . . . , p_C \right \}$). The loss function is given by:
\begin{align}
\label{equ:imba} 
L_{\text{cls}}(p,u) = -\log p_u
\end{align}
Lin et al. proposed a novel focal loss\cite{lin2017focal} which suppressed signals from easy samples. Instead of discarding all easy samples, they assigned an importance weight to each sample w.r.t its loss value as: 
\begin{align}
 L_{\text{FL}} = -\alpha(1-p_u)^\gamma\log(p_u)
 \label{equ:fl}
\end{align} 
where $\alpha$ and $\gamma$ were parameters to control the importance weight. The gradient signals of easy samples got suppressed which led the training process to focus more on hard proposals. Li et al.~\cite{li2018gradient} adopt a similar idea from focal loss and propose a novel gradient harmonizing mechanism (GHM). The new proposed GHM not only suppressed easy proposals but also avoided negative impact of outliers. Shrivastava et al.~\cite{shrivastava2016training} proposed an online hard example mining strategy which was based on a similar principle as Liu et al.'s SSD ~\cite{liu2016ssd} to automatically select hard examples for training. Different from Liu et al., online hard negative mining only considered difficulty information but ignored categorical information, which meant the ratio of foreground and background was not fixed in each mini-batch. They argued that difficult samples played a more important role than class imbalance in object detection task.

\subsubsection{Localization Refinement}
\begin{figure}[t]
\begin{center}
\includegraphics[width=0.8\linewidth]{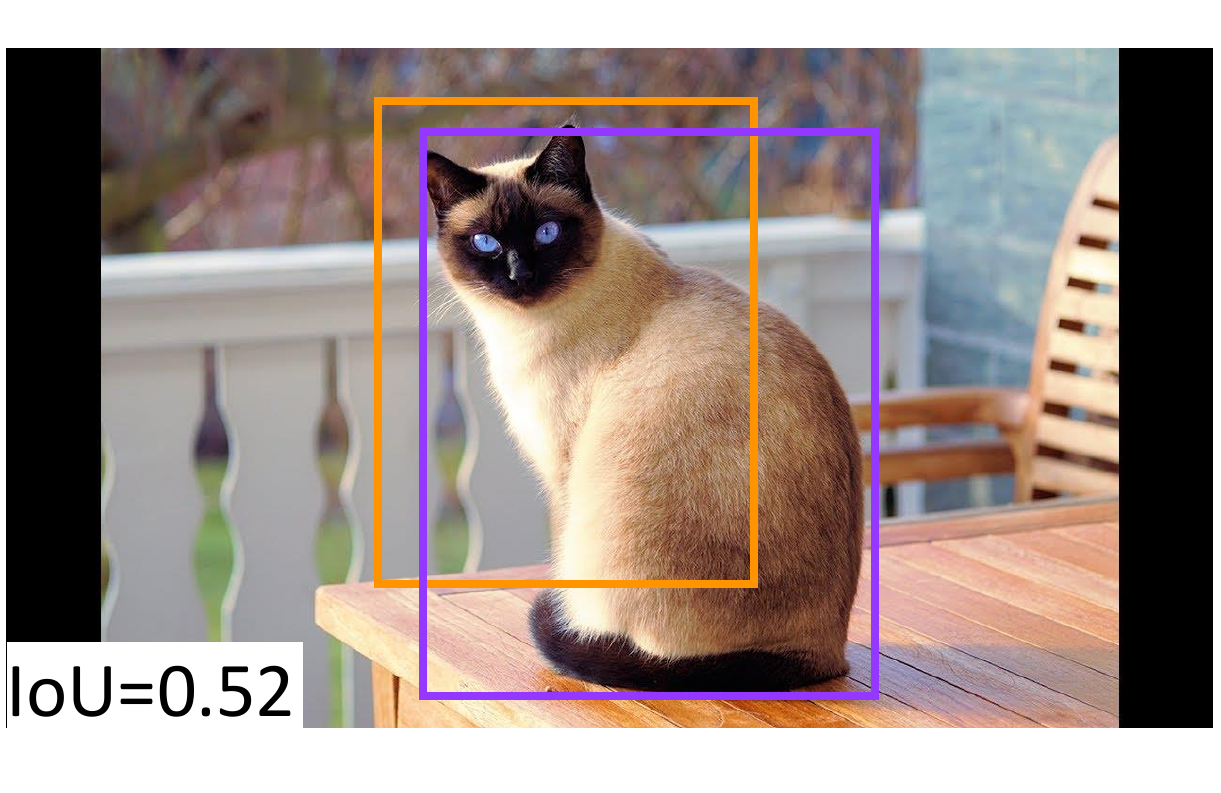}
\end{center}
\vspace{-0.3mm}
\caption{Example of failure case of detection in high IoU threshold. Purple box is ground truth and yellow box is prediction. In low IoU requirement scenario, this prediction is correct while in high IoU threshold, it's a false positive due to insufficient overlap with objects.}
\label{fig:high}
\end{figure}

An object detector must provide a tight localization prediction (bbox or mask) for each object. To do this, many efforts refine the preliminary proposal prediction to improve the localization. Precise localization is challenging because predictions are commonly focused on the most discriminative part of the objects, and not necessarily the region containing the object. In some scenarios, the detection algorithms are required to make high quality predictions (high IoU threshold)
See Fig. \ref{fig:high} for an illustration of how a detector may fail in a high IoU threshold regime. A general approach for localization refinement is to generate high quality proposals (See Sec \ref{sec:prop}). In this section, we will review some other methods for localization refinement. In R-CNN framework, the L-2 auxiliary bounding box regressors were learned to refine localizations, and in Fast R-CNN, the smooth L1 regressors were learned via an end-to-end training scheme as:
\begin{align}
 L_{\text{reg}}(t^c, v) = \sum_{i \in \left \{ x,y,w,h \right \}} \text{SmoothL1}(t^c_i-v_i)
 \label{equ:reg1}
\end{align}
\begin{align}
 \text{SmoothL1}(x) = \left\{\begin{matrix}
0.5x^2 & \text{if } \left | x \right |<1 \\
\left | x \right | -0.5 & \text{otherwise}
\end{matrix}\right.
 \label{equ:reg2}
\end{align}
where the predicted offset is given by $t^c = (t^c_x,t^c_y,t^c_w,t^c_h)$ for each target class, and $v$ denotes ground truth of object bounding boxes($v = (v_x, v_y, v_w, v_h)$). $x, y, w, h$ denote bounding box center, width and height respectively.

Beyond the default localization refinement, some methods learn auxiliary models to further refine localizations. Gidaris et al.~\cite{gidaris2015object} introduced an iterative bounding box regression method, where an R-CNN was applied to refine learned predictions. Here the predictions were refined multiple times. 
Gidaris et al.~\cite{gidaris2016locnet} proposed LocNet which modeled the distribution of each bounding box and refined the learned predictions. Both these approaches required a separate component in the detection pipeline, and prevent joint optimization. 

Some other efforts ~\cite{zagoruyko2016multipath,lu2019grid} focus on designing a unified framework with modified objective functions. In MultiPath Network, Zagoruyko et al. ~\cite{zagoruyko2016multipath} developed an ensemble of classifiers which were optimized with an integral loss targeting various quality metrics. Each classifier was optimized for a specific IoU threshold and the final prediction results were merged from these classifiers. Tychsen et al. proposed Fitness-NMS~\cite{tychsen2017improving} which learned novel fitness score function of IoU between proposals and objects. They argued that existing detectors aimed to find \textit{qualified} predictions instead of \textit{best} predictions and thus highly quality and low quality proposals received equal importance. Fitness-IoU assigned higher importance to highly overlapped proposals. They also derived a bounding box regression loss based on a set of IoU upper bounds to maximum the IoU of predictions with objects. Inspired by CornerNet~\cite{law2018cornernet} and DeNet~\cite{tychsen2017denet}, Lu et al.~\cite{lu2019grid} proposed a Grid R-CNN which replaced linear bounding box regressor with the principle of locating corner keypoints corner-based mechanism. 

\subsubsection{Cascade Learning}
Cascade learning is a coarse-to-fine learning strategy which collects information from the output of the given classifiers to build stronger classifiers in a cascaded manner. Cascade learning strategy was first used by Viola and Jones~\cite{viola2001rapid} to train the robust face detectors. In their models, a lightweight detector first rejects the majority easy negatives and feeds hard proposals to train detectors in next stage. For deep learning based detection algorithms, Yang et al.~\cite{yang2016craft} proposed CRAFT (Cascade Region-proposal-network And FasT-rcnn) which learned RPN and region classifiers with a cascaded learning strategy. 
CRAFTS first learned a standard RPN followed by a two-class Fast RCNN which rejected the majority easy negatives. The remaining samples were used to build the cascade region classifiers which consisted of two Fast RCNNs. 
Yang et al.~\cite{yang2016exploit} introduced layer-wise cascade classifiers for different scale objects in different layers. Multiple classifiers were placed on different feature maps and classifiers on shallower layers would reject easy negatives. The remaining samples would be fed into deeper layers for classification. RefineDet~\cite{zhang2018single} and Cascade R-CNN~\cite{cai2018cascade} utilized cascade learning methods in refining object locations. They built multi-stage bounding box regressors and bounding box predictions were refined in each stage trained with different quality metrics. 
Cheng et al.~\cite{cheng2018revisiting} observed the failure cases of Faster RCNN, and noticed that even though the localization of objects was good, there were several classification errors. They attributed this to sub-optimal feature representation due to sharing of features and joint multi-task optimization, for classification and regression; and they also argued that the large receptive field of Faster RCNN induce too much noise in the detection process. 
They found that vanilla RCNN was robust to these issues. Thus, they built a cascade detection system based on Faster RCNN and RCNN to complement each other. Specifically, A set of initial predictions were obtained from a well trained Faster RCNN, and these predictions were used to train RCNN to refine the results.

\subsubsection{Others}
There are some other learning strategies which offer interesting directions, but have not yet been extensively explored. We split these approaches into four categories: adversarial learning, training from scratch and knowledge distillation.

\textbf{Adversarial Learning.} Adversarial learning has shown significant advances in generative models. The most famous work applying adversarial learning is generative adversarial network(GAN)~\cite{goodfellow2014generative} where a generator is competing with a discriminator. The generator tries to model data distribution by generating fake images using a noise vector input and use these fake images to confuse the discriminator, while the discriminator competes with the generator to identify the real images from fake images. GAN and its variants~\cite{CycleGAN2017,radford2015unsupervised,brock2018large} have shown effectiveness in many domains and have also found applications in object detection. Li et al.~\cite{li2017perceptual} proposed a new framework Perceptual GAN for small object detection. 
The learnable generator learned high-resolution feature representations of small objects via an adversarial scheme. Specifically, its generator learned to transfer low-resolution small region features into high-resolution features and competed with the discriminator which identified real high-resolution features.
Finally the generator learned to generate high quality features for small objects. Wang et al.~\cite{wang2017fast} proposed A-Fast-R-CNN which was trained by generated adversarial examples. They argued the difficult samples were on long tail so they introduced two novel blocks which automatically generated features with occlusion and deformation. Specifically, a learned mask was generated on region features followed by region classifiers. In this case, the detectors could receive more adversarial examples and thus become more robust.

\textbf{Training from Scratch.} Modern object detectors heavily rely on pre-trained classification models on ImageNet, however, the bias of loss functions and data distribution between classification and detection can have an adversarial impact on the performance. Finetuning on detection task can relieve this issue, but cannot fully get rid of the bias. Besides, transferring a classification model for
detection in a new domain can lead to more challenges (from RGB to MRI data etc.). 
Due to these reasons, there is a need to train detectors from scratch, instead of relying on pretrained models. 
The main difficulty of training detectors from scratch is the training data of object detection is often insufficient and may lead to overfitting. Different from image classification, object detection requires bounding box level annotations and thus, annotating a large scale detection dataset requires much more effort and time(ImageNet has 1000 categories for image classification while only 200 of them have detection annotations).

There are some works~\cite{shen2017dsod,he2018Rethinking,zhu2019scratchdet} exploring training object detectors
from scratch. 
Shen et al.~\cite{shen2017dsod} first proposed a novel framework DSOD (Deeply Supervised Object Detectors) to train detectors from scratch. They argued deep supervision with a densely connected network structure could significantly reduce
optimization difficulties. Based on DSOD, Shen et al.~\cite{shen2017learning} proposed a gated recurrent feature pyramid which dynamically adjusted supervision intensities of intermediate layers for objects with different scales. They defined a recurrent feature pyramid structure to squeeze both spatial and semantic information into a single prediction layer, which further reduced parameter numbers leading to faster convergence. In addition, the gate-control structure on feature pyramids adaptively adjusted the supervision at different scales based on the size of objects. Their method was more powerful than original DSOD. However, later He et al.~\cite{he2018Rethinking} validated the difficulty of training detectors from scratch on MSCOCO and found that the vanilla detectors could obtain a competitive performance with at least 10K annotated images. Their findings proved no specific structure was required for training from scratch which contradicted the previous work. 

\textbf{Knowledge Distillation.} Knowledge distillation is a training strategy which distills the knowledge in an ensemble of models into a single model via teacher-student training scheme. This learning strategy was first used in image classification~\cite{hinton2015distilling}. In object detection, some works~\cite{Li2017Mimicking,cheng2018revisiting} also investigate this training scheme to improve detection performance. Li et al.~\cite{Li2017Mimicking} proposed a light weight detector whose optimization was carefully guided by a heavy but powerful detector. This light detector could achieve comparable detection accuracy by distilling knowledge from the heavy one, meanwhile having faster inference speed. Cheng et al.~\cite{cheng2018revisiting} proposed a Faster R-CNN based detector which was optimized via teacher-student training scheme. An R-CNN model is used as teacher network to guide the training process. Their framework showed improvement in detection accuracy compared with traditional single model optimization strategy.

\subsection{Testing Stage}
Object detection algorithms make a dense set of predictions and thus these predictions cannot be directly used for evaluation due to heavy duplication. In addition, some other learning strategies are required to further improve the detection accuracy. These strategies improve the quality of prediction or accelerate the inference speed. In this section, we introduce these strategies in testing stage including duplicate removal, model acceleration and other effective techniques.

\subsubsection{Duplicate Removal}
\begin{figure}
\centering
\includegraphics[width=3.5in]{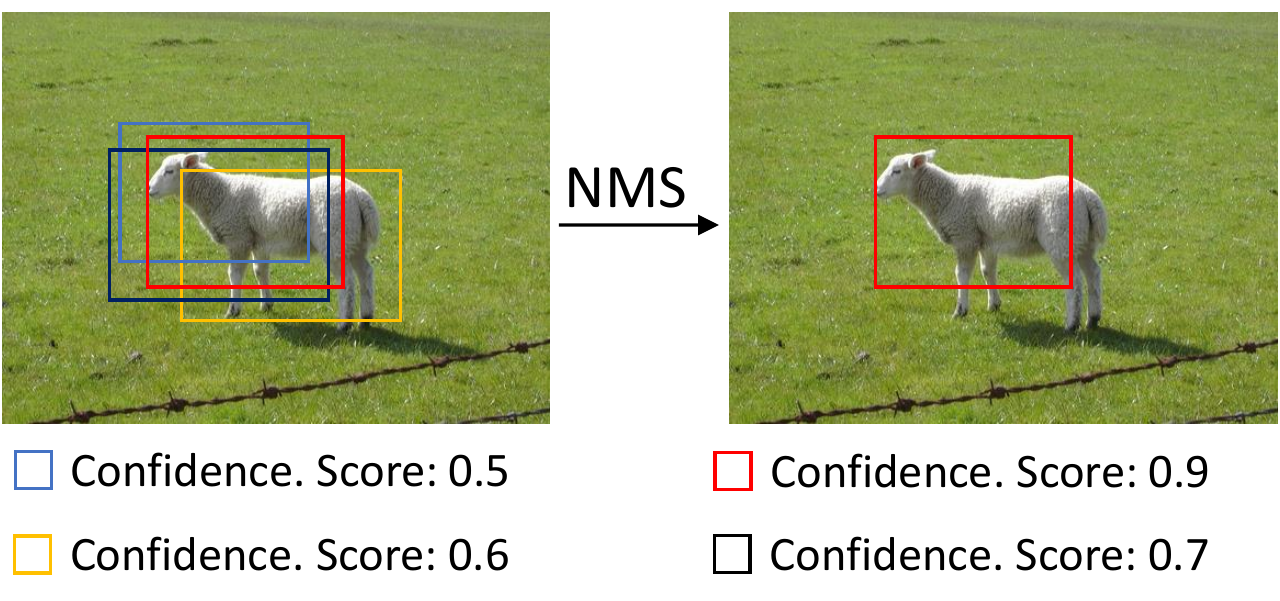}
\caption{Duplicate predictions are eliminated by NMS operation. The most-confident box is kept, and all other boxes surrounding it will be removed.}
\label{fig:nms}
\end{figure}

Non maximum suppression(NMS) is an integral part of object detection to remove duplicate false positive predictions (See Figure \ref{fig:nms}). Object detection algorithms make a dense set of predictions with several duplicate predictions. For one-stage detection algorithms which generate a dense set of candidate proposals such as SSD~\cite{liu2016ssd} or DSSD (Deconvolutional Single Shot Detector)~\cite{fu2017dssd}, the proposals surrounding the same object may have similar confidence scores, leading to false positives. For two-stage detection algorithms which generates a sparse set of proposals, the bounding box regressors will pull these proposals close to the same object and thus lead to the same problem. The duplicate predictions are regarded as false positives and will receive penalties in evaluation, so NMS is needed to remove these duplicate predictions. Specifically, for each category, the prediction boxes are sorted according to the confidence score and the box with highest score is selected. This box is denoted as $M$. Then IoU of other boxes with $M$ is calculated, and if the IoU value is larger than a predefined threshold $\Omega_{\text{test}}$, these boxes will are removed. This process is repeated for all remaining predictions. More formally, the confidence score of box $B$ which overlaps with $M$ larger than $\Omega_{\text{test}}$ will be set to zero:
\begin{align}
  \text{Score}_B = \left\{\begin{matrix}
  \text{Score}_B& \text{IoU}(B, M)<\Omega_{\text{test}}\\
  0 & \text{IoU}(B, M) \ge \Omega_{\text{test}}
  \end{matrix}\right.
\end{align}
However, if an object just lies within $\Omega_{\text{test}}$ of $M$, NMS will result in a missing prediction, and this scenario is very common in clustered object detection. Navaneeth et al.~\cite{Navaneeth2017soft} introduced a new algorithm Soft-NMS to address this issue. Instead of directly eliminating the prediction $B$, Soft-NMS decayed the confidence score of $B$ as a continuous function $F$($F$ can be linear function or guassian function) of its overlaps with $M$. This is given by:
\begin{align}
  \text{Score}_B = \left\{\begin{matrix}
  \text{Score}_B& \text{IoU}(B, M)<\Omega_{\text{test}}\\
  F(\text{IoU}(B, M)) & \text{IoU}(B, M) \ge \Omega_{\text{test}}
  \end{matrix}\right.
\end{align}
Soft-NMS avoided eliminating prediction of clustered objects and showed improvement in many common benchmarks. Hosong et
al.\cite{Hosang2017Learning} introduced a network architecture designed to perform NMS based on confidence scores and
bounding boxes, which was optimized separately from detector training in a supervised way. They argued the reason for duplicate
predictions was that the detector deliberately encouraged multiple high score detections per object instead of rewarding one high
score. Based on this, they designed the network following two motivations: (i) a loss penalizing double detections to
push detectors to predict exactly one precise detection per object; (ii) joint processing of detections nearby to give the detector
information whether an object is detected more than once. The new proposed model did not discard detections but instead
reformulated NMS as a re-scoring task that sought to decrease the score of detections that cover objects that already have been
detected.

\subsubsection{Model Acceleration}
Application of object detection for real world application requires the algorithms to function in an efficient manner. Thus, evaluating detectors on efficiency metrics is important. Although current state-of-the-art algorithms~\cite{huang2017speed,he2016deep} can achieve very strong results on public datasets, their inference speeds make it difficult to apply them into real applications. In this section we review several works on accelerating detectors. 
Two-stage detectors are usually slower than one-stage detectors because they have two stages - one proposal generation and one region classification, which makes them computationally more time consuming than one-stage detectors which directly use one network for both proposal generation and region classification. 
R-FCN \cite{dai2016r} built spatially-sensitive feature maps and extracted features with position sensitive ROI Pooling to share computation costs. However, the number of channels of spatially-sensitive feature maps significantly increased with the number of categories. Li et al.~\cite{li2017light} proposed a new framework Light Head R-CNN which significantly reduced the number of channels in the final feature map (from 1024 to 16) instead of sharing all computation. Thus, though computation was not shared across regions, but the cost could be neglected. 

From the aspect of backbone architecture, a major computation cost in object detection is feature extraction~\cite{ren2015faster}. A simple idea to accelerate detection speed is to replace the detection backbone with a more efficient backbone, e.g.,  MobileNet~\cite{howard2017mobilenets,sandler2018inverted} was an efficient CNN model with depth-wise convolution layers which was also adopted into many works such as \cite{wong2018tiny} and \cite{li2018tiny}. PVANet~\cite{kim2016pvanet} was proposed as a new network structure with CReLu~\cite{shang2016under} layer to reduce non-linear computation and accelerated inference speed. Another approach is to optimize models off-line, such as model compression and quantization ~\cite{Kim2015Compression,He2017Channel,Gong2014Compressing,lin2017deep,Wu2016Quantized,Han2015Deep,han2015learning} on the learned models. Finally, NVIDIA Corporation\footnote{\url{https://www.nvidia.com/en-us/}} released an acceleration toolkit TensorRT\footnote{\url{https://developer.nvidia.com/tensorrt}} which optimized the computation of learned models for deployment and thus significantly sped up the inference. 

\subsubsection{Others}
Other learning strategies in testing stage mainly comprise the transformation of input image to improve the detection accuracy. Image pyramids~\cite{he2016deep,zhang2018single} are a widely used technique to improve detection results, which build a hierarchical image set at different scales and make predictions on all of these images. The final detection results are merged from the predictions of each image. Zhang et al.~\cite{Zhang_2017_ICCV,zhang2018single} used a more extensive image pyramid structure to handle different scale objects. They resized the testing image to different scales and each scale was responsible for a certain scale range of objects. Horizontal Flipping~\cite{he2017mask,zhang2018single} was also used in
the testing stage and also showed improvement. These learning strategies largely improved the the capability of detector to handle different scale objects and thus were widely used in public detection competitions. However, they also increase computation cost and thus were not suitable for real world applications.

\section{Applications}\label{sec:app}
Object detection is a fundamental computer vision task and there are many real world applications based on this task. Different from generic object detection, these real world applications commonly have their own specific properties and thus carefully-designed detection algorithms are required. In this section, we will introduce several real world applications such as face detection and pedestrian detection. 

\subsection{Face Detection}
Face detection is a classical computer vision problem to detect human faces in the images, which is often the first step towards many real-world applications with human beings, such as face verification, face alignment and face recognition. There are some critical differences between face detection and generic detection: i) the range of scale for objects in face detection is much larger than objects in generic detection. Moreover occlusion and blurred cases are more common in face detection; ii) Face objects contain strong structural information, and there is only one target category in face detection. Considering these properties of face detection, directly applying generic detection algorithms is not an optimal solution as there could be some priors that can exploited to improve face detection.

In early stages of research before the deep learning era, face detection~\cite{viola2004robust,osuna1997training,ratsch2004efficient,romdhani2001computationally} was mainly based on sliding windows, and dense image grids were encoded by hand-crafted features followed by training classifiers to find and locate objects. Notably, Viola and Jones~\cite{viola2004robust} proposed a pioneering cascaded classifiers using AdaBoost with Haar features for face detection and obtained excellent performance with high real time prediction speed. After the progresses of deep learning in image classification, face detectors based on deep learning significantly outperformed traditional face detectors~\cite{sun2018face,liu2017multi,tang2018pyramidbox,chi2018selective,li2019dsfd}.

Current face detection algorithms based on deep learning are mainly extended from generic detection frameworks such as Fast
R-CNN and SSD. These algorithms focus more on learning robust feature representations. In order to handle extreme scale variance, multi-scale feature learning methods discussed before have been widely used in face detection. Sun et al.~\cite{sun2018face} proposed a Fast R-CNN based framework which integrated multi-scale features for prediction and converted the resulting detection bounding boxes into ellipses as the regions of human faces are more elliptical than rectangular. Zhang et al.~\cite{Zhang_2017_ICCV} proposed one-stage S3FD which found faces on different feature maps to detect faces at a large range of scales. They made predictions on larger feature maps to capture small-scale face information. Notably, a set of anchors were carefully designed according to empirical receptive fields and thus provided a better match to the faces. Based on S3FD, Zhang et al. \cite{zhang2017feature} proposed a novel network structure to capture multi-scale features in different stages. The new proposed feature agglomerate structure integrated features at different scales in a hierarchical way. Moreover, a hierarchical loss was proposed to reduce the training difficulties. Single Stage Headless Face Detector (SSH)~\cite{najibi2017ssh} was another one-stage face detector which combined different scale features for prediction. Hu et al.~\cite{Hu_2017_CVPR} gave a detailed analysis of small face detection and proposed a light weight face detector consisting of multiple RPNs, each of which was responsible for a certain range of scales. Their method could effectively handle face scale variance but it was slow for real world usage. Unlike this method, Hao et al.~\cite{hao2017scale} proposed a Scale Aware Face network which addresses scale issues without incurring significant computation costs. 
They learned a scale aware network which modeled the scale distribution of faces in a given image and guided zoom-in or zoom-out operations to make sure that the faces were in desirable scale. The resized image was fed into a single scale light weight face detector. 
Wang et al.~\cite{wang2017face} followed RetinaNet~\cite{lin2017focal} and utilized more dense anchors to handle faces in a large range of scales. Moreover, they proposed an attention function to account for context information, and to highlight the discriminative features. 
Zhang et al.~\cite{zhang2016joint} proposed a deep cascaded multi-task face detector with cascaded structure (MTCNN). MTCNN had three stages of carefully designed CNN models to predict faces in a coarse-to-fine style. Further, they also proposed a new online hard negative mining strategy to improve the result. Samangouei et al.~\cite{samangouei2018face} proposed a Face MegNet which allowed information flow of small faces without any skip connections by placing a set of deconvolution layers before RPN and ROI Pooling to build up finer face representations.

In addition to multi-scale feature learning, some frameworks were focused on contextual information. Face objects have strong physical relationships with the surrounding contexts (commonly appearing with human bodies) and thus encoding contextual information became an effective way to improve detection accuracy. Zhang et al.~\cite{zhang2018face} proposed FDNet based on ResNet with larger deformable convolutional kernels to capture image context. Zhu et al.~\cite{zhu2017cms} proposed a Contextual Multi-Scale Region-based Convolution Neural Network (CMS-RCNN) in which multi-scale information was grouped both in region proposal and ROI detection to deal with faces at various range of scale. In addition, contextual information around faces is also considered in training detectors. Notably, Tang et al.~\cite{tang2018pyramidbox} proposed a state-of-the-art context assisted single shot face detector, named PyramidBox to handle the hard face detection problem. Observing the importance of the context, they improved the utilization of contextual information in the following three aspects: i) first, a novel context anchor was designed to supervise high-level contextual feature learning by a semi-supervised method, dubbed as PyramidAnchors; ii) the Low-level Feature Pyramid Network was developed to combine adequate high-level context semantic features and low-level facial features together, which also allowed the PyramidBox to predict faces at all scales in a single shot; and iii) they introduced a context sensitive structure to increase the capacity of prediction network to improve the final accuracy of output. In addition, they used the method of data-anchor-sampling to augment the training samples across different scales, which increased the diversity of training data for smaller faces. Yu et al.\cite{yu2018anchor} introduced a context pyramid maxout mechanism to explore image contexts and devised an efficient anchor based cascade framework for face detection which optimized anchor-based detector in cascaded manner. Zhang et al.~\cite{zhang2017detecting} proposed a two-stream contextual CNN to adaptively capture body part information. In addition, they proposed to filter easy non-face regions in the shallow layers and leave difficult samples to deeper layers.

Beyond efforts on designing scale-robust or context-assistant detectors, Wang et al.~\cite{wang2017face} developed a framework from the perspective of loss function design. Based on vanilla Faster R-CNN framework, they replaced original softmax loss with a center loss which encouraged detectors to reduce the large intra-class variance in face detection. They explored multiple technologies in improving Faster R-CNN such as fixed-ratio online hard negative mining, multi-scale training and multi-scale testing, which made vanilla Faster R-CNN adaptable to face detection. Later, Wang et al.~\cite{wang2017detecting} proposed Face R-FCN which was based on vanilla R-FCN. Face R-FCN distinguished the contribution of different facial parts and introduced a novel position-sensitive average pooling to re-weight the response on final score maps. This method achieved state-of-the-art results on many public benchmarks such as FDDB~\cite{fddbTech} and WIDER FACE\cite{yang2016wider}.

\subsection{Pedestrian Detection}
Pedestrian detection is an essential and significant task in any intelligent video surveillance system. Different from generic
object detection, there are some properties of pedestrian detection different from generic object detection: 
i) Pedestrian objects are well structured objects with nearly fixed aspect ratios (about 1.5), but they also lie at a large range of scales;
ii) Pedestrian detection is a real world application, and hence the challenges such as crowding, occlusion and blurring are commonly exhibited. For example, in the CityPersons dataset, there are a total of 3157 pedestrian annotations in the validation subset, among which 48.8\% overlap with another annotated pedestrian with Intersection over Union (IoU) above 0.1. Moreover, 26.4\% of all
pedestrians have considerable overlap with another annotated pedestrian with IoU above 0.3. The highly frequent crowd occlusion harms the performance of pedestrian detectors; 
iii) There are more hard negative samples (such as traffic light, Mailbox etc.) in pedestrian detection due to complicated contexts. 

Before the deep learning era, pedestrian detection algorithms ~\cite{dalal2005histograms,han2014local,dollar2009integral,dollar2014fast,papageorgiou1998general}
were mainly extended from Viola Jones frameworks~\cite{viola2004robust} by exploiting Integral Channel Features with a sliding window strategy to locate objects, followed by region classifiers such as SVMs. The early works were mainly focused on designing robust feature descriptors for classification. For example, Dalal and Triggs~\cite{dalal2005histograms} proposed the histograms of oriented gradient (HOG) descriptors, while Paisitkriangkrai et al.~\cite{papageorgiou1998general} designed a feature descriptor based on low-level visual cues and spatial pooling features. These methods show promising results on pedestrian detection benchmarks but were mainly based on hand-crafted features.

Deep learning based methods for pedestrian detection ~\cite{li2018scale,hosang2015taking,angelova2015real,brazil2017illuminating,du2017fused,wang2018pcn,xu2017learning,benenson2014ten,cai2015learning,sermanet2013pedestrian} showed excellent performance and achieved state-of-the-art results on public benchmarks. Angelova et al~\cite{angelova2015real} proposed a real-time pedestrian detection framework using a cascade of deep convolutional networks. In their work, a large number of easy negatives were rejected by a tiny model and the remaining hard proposals were then classified by a large deep networks. 
Zhang et al.~\cite{zhang2016faster} proposed decision tree based framework. In their method, multiscale feature maps were used to extract pedestrian features, which were later fed into boosted decision trees for classification. Compared with FC layers, boosted decision trees applied bootstrapping strategy for mining hard negative samples and achieved better performance. 
Also to reduce the impact of large variance in scales, Li et al.~\cite{li2018scale} proposed Scale-aware Fast R-CNN (SAF RCNN) which inserted multiple built-in networks into the whole detection framework. The proposed SAF RCNN detected different scale pedestrian instances using different sub-net.
Further, Yang et al.~\cite{yang2016exploit} inserted Scale Dependent Pooling (SDP) and Cascaded Rejection Classifiers (CRC) into Fast RCNN to handle pedestrian scale issue. According to the height of the instances, SDP extracted region features from suitable scale feature maps, while CRC rejected easy negative samples in shallower layers. 
Wang et al.~\cite{wang2018repulsion} proposed a novel Repulsion Loss to detect pedestrians in a crowd. They argued that detecting a pedestrian in a crowd made it very sensitive to the NMS threshold, which led to more false positives and missing objects. The new proposed repulsion loss pushed the proposals into their target objects but also pulled them away from other objects and their target proposals. Based on their idea, 
Zhang et al.~\cite{zhang2018occlusion} proposed an Occlusion-aware R-CNN (OR-CNN) which was optimized by Aggression Loss. The new loss function encouraged the proposals to be close to the objects and other proposals with the same targeted proposals. 
Mao et al.~\cite{mao2017can} claimed that properly aggregating extra features into pedestrian detector can boost detection accuracy. In their paper, they explored different kinds of extra features useful in improving accuracy and proposed a new method to use these features. The new proposed component HyperLearner aggregated extra features into vanilla DCNN detector via jointly optimization fashion and no extra input was required in inference stage.

For pedestrian detection, one of the most significant challenges is to handle occlusion~\cite{tian2015deep,ouyang2013joint,mathias2013handling,ouyang2013modeling,duan2010structural,enzweiler2010multi,zhou2018bi,ouyang2012discriminative,tang2014detection,ouyang2013single,zhang2018occlusion,shet2007bilattice}. A straightforward method is to use part-based models which learn a series of part detectors and integrate the results of part detectors to locate and classify objects.
Tian et al.~\cite{tian2015deep} proposed DeepParts which consisted of multiple parts detectors. During training, the important body parts were automatically selected from a part pool covered all scale parts of the body, and for each selected part, a detector was learned to handle occlusions.
To avoid integrate inaccurate scores of part models, Ouyang and Wang~\cite{ouyang2012discriminative} proposed a framework which modeled visible parts as hidden variables in training the models. In their paper, the visible relationship of overlapping parts were learned by a discriminative deep models, instead of being manually defined or even being assumed independent. 
Later, Ouyang et al.~\cite{ouyang2013single} addressed this issue from another aspect. They proposed a mixture network to capture unique visual information which was formed by crowded pedestrians. To enhance the final predictions of single-pedestrian detectors, a probabilistic framework was learned to model the relationship between the configurations estimated by single- and multi-pedestrian detectors.  
Zhang et al.~\cite{zhang2018occlusion} proposed an occlusion-aware ROI Pooling layer which integrated the prior structure information of pedestrian with visibility prediction into the final feature representations. The original region was divided into five parts and for each part a sub-network enhanced the original region feature via a learned visibility score for better representations.
Zhou et al.~\cite{zhou2018bi} proposed Bi-box which simultaneously estimated pedestrian detection as well as visible parts by regressing two bounding boxes, one for the full body and the other for visible part. In addition, a new positive-instance sampling criterion was proposed to bias positive training examples with large visible area, which showed effectiveness in training occlusion-aware detectors.

\subsection{Others}
There are some other real applications with object detection techniques, such as logo detection and video object detection. 

Logo detection is an important research topic in e-commerce systems. Compared to generic detection, logo instance is much smaller with strong non-rigid transformation. Further, there are few logo detection baselines available. To address this issue, Su et al.~\cite{su2018scalable} adopted the learning principle of webly data learning which automatically mined information from noisy web images and learns models with limited annotated data. Su et al.~\cite{su2017deep} described an image synthesising method to successfully learn a detector with limited logo instances. Hoi et al.~\cite{hoi2015logo} collected a large scale logo dataset from an e-commerce website and conducted a comprehensive analysis on the problem logo detection.

Existing detection algorithms are mainly designed for still images and are suboptimal for directly applying in videos for object detection. To detect objects in videos, there are two major differences from generic detection: temporal and contextual information. The location and appearance of objects in video should be temporally consistent between adjacent frames. Moreover, a video consists of hundreds of frames and thus contains far richer contextual information compared to a single still image. Han et al.~\cite{han2016seq} proposed a Seq-NMS which associates detection results of still images into sequences. Boxes of the same sequence are re-scored to the average score across frames, and other boxes along the sequence are suppressed by NMS. Kang et al. proposed Tubelets with Convolutional Neural Networks (T-CNN)~\cite{kang2016object} which was extended from Faster RCNN and incorporated the temporal and contextual information from tubelets(box sequence over time). T-CNN propagated the detection results to the adjacent frames by optical flow, and generated tubelets by applying tracking algorithms from high-confidence bounding boxes. The boxes along the tubelets were re-scored based on tubelets classification. 

There are also many other real-world applications based on object detection such as vehicle detection~\cite{zhou2016dave,gebru2017fine,majid2018shuffledet}, traffic-sign detection~\cite{zhu2016traffic,pon2018hierarchical} and skeleton detection~\cite{ke2017srn,shen2016object}.

\section{Detection Benchmarks}
\label{sec:bench}
In this section we will show some common benchmarks of generic object detection, face detection and pedestrian detection. We will first present some widely used datasets for each task and then introduce the evaluation
metrics. 

\begin{figure*}[htb]
\centering
\includegraphics[width=7.2in]{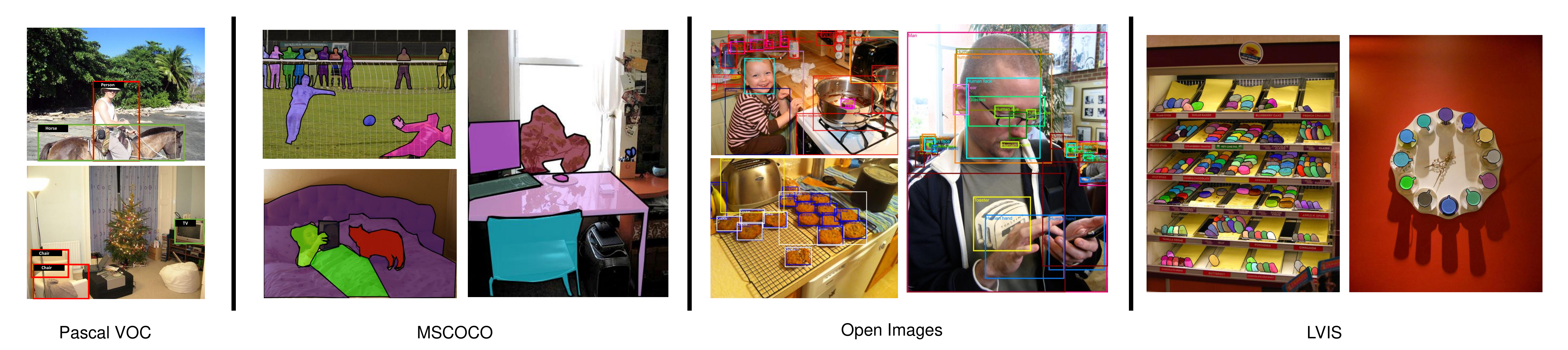}
\caption{ Some examples of Pascal VOC, MSCOCO, Open Images and LVIS.}
\label{fig:challenge}
\end{figure*}

\subsection{Generic Detection Benchmarks}
\noindent{\bf Pascal VOC2007~\cite{Everingham10}} is a mid scale dataset for object detection with 20 categories. There are
three image splits in VOC2007: training, validation and test with 2501, 2510 and 5011 images respectively. 

\noindent{\bf Pascal VOC2012~\cite{Everingham10}} is a mid scale dataset for object detection which shares the same 20
categories with Pascal VOC2007. There are three image splits in VOC2012: training, validation and test with 5717, 5823 and
10991 images respectively. The annotation information of VOC2012 test set is not available. 

\noindent{\bf MSCOCO~\cite{lin2014microsoft}} is a large scale dataset for with 80 categories. There are three image splits in
MSCOCO: training, validation and test with 118287, 5000 and 40670 images respectively. The annotation information of
MSCOCO test set is not available. 

\noindent{\bf Open Images~\cite{kuznetsova2018open}} contains 1.9M images with 15M objects of 600 categories. The 500 most frequent categories are used to evaluate detection benchmarks, and more than 70\% of these categories have over 1000 training samples.

\noindent{\bf LVIS~\cite{Gupta_2019_CVPR}} is a new collected benchmark with 164000 images and 1000+ categories. It's a new dataset without any existing results so we leave the details of LVIS in future work section (Section \ref{sec:conc}). 

\noindent{\bf ImageNet~\cite{deng2009imagenet}} is also a important dataset with 200 categories. However, the scale of ImageNet is huge and the object scale range is similar to VOC datasets, so it is not a commonly used benchmarks for detection algorithms.

\noindent{\bf Evaluation Metrics:} The detail of evaluation metrics are shown in Tab. \ref{tab:eval}, both detection accuracy and inference speed are used to evaluate detection algorithms. For detection accuracy, mean Average Precision(mAP) is used as evaluation metric for all these challenges. For VOC2012, VOC2007 and ImageNet, IoU threshold of mAP is set to 0.5, and for MSCOCO, more comprehensive evaluation metrics are applied. There are six evaluation scores which demonstrates different capability of detection algorithms, including performance on different IoU thresholds and on different scale objects. Some examples of listed datasets(Pascal VOC, MSCOCO, Open Images and LVIS) are shown in Fig. \ref{fig:challenge}.

\begin{table}[!t]
\centering
\renewcommand{\arraystretch}{1.2}
\setlength\arrayrulewidth{0.2mm}
\setlength\tabcolsep{1pt}
\resizebox*{9.cm}{!}{
\begin{tabular}{!{\vrule width1.2bp}c|c|l|l!{\vrule width1.2bp}}
\Xhline{1pt}
\bf Alias & \bf Meaning & \multicolumn{2}{c!{\vrule width1.2bp}}{\bf Definition and Description} \\
\hline
FPS & \raisebox{0.1ex}[0pt]{ \shortstack [c] {Frame per second}} & \multicolumn{2}{l!{\vrule width1.2bp}}{\shortstack [l] {The number of images processed per second.}} \\
\hline
$\Omega$ & \raisebox{0.1ex}[0pt]{ \shortstack [c] {IoU threshold}} & \multicolumn{2}{l!{\vrule width1.2bp}}{\shortstack [l] {The IoU threshold to evaluate localization.}} \\
\hline
D$_\gamma$ & \raisebox{0.1ex}[0pt]{ \shortstack [c] {All Predictions}} & \multicolumn{2}{l!{\vrule width1.2bp}}{\shortstack [l] {Top $\gamma$ predictions returned by the detectors \\with highest confidence score.}} \\
\hline
TP$_\gamma$ & \raisebox{0.1ex}[0pt]{ \shortstack [c] {True Positive}} & \multicolumn{2}{l!{\vrule width1.2bp}}{\shortstack [l] {Correct predictions from sampled predictions }} \\
\hline
FP$_\gamma$ & \raisebox{0.1ex}[0pt]{ \shortstack [c] {False Positive}} & \multicolumn{2}{l!{\vrule width1.2bp}}{\shortstack [l] {False predictions from sampled predictions.}} \\
\hline
P$_\gamma$ & \raisebox{0.3ex}[0pt]{ \shortstack [c] {Precision}} & \multicolumn{2}{l!{\vrule width1.2bp}}{ \shortstack [l] {The fraction of TP$_\gamma$ out of D$_\gamma$.}} \\
\hline
R$_\gamma$ & \raisebox{0.3ex}[0pt]{\shortstack [c] {Recall}} & \multicolumn{2}{l!{\vrule width1.2bp}}{\shortstack [l] {The fraction of TP$_\gamma$ out of all positive samples.}} \\
\hline
 AP & \shortstack [c] {Average Precision} & \multicolumn{2}{l!{\vrule width1.2bp}}{\shortstack [l] {Computed over the different levels of recall by varying the $\gamma$.}} \\
\hline
 mAP & \shortstack [c] {mean AP} & \multicolumn{2}{l!{\vrule width1.2bp}}{\shortstack [l] {Average score of AP across all classes.}} \\
\hline
TPR & \shortstack [c] {True Positive Rate} & \multicolumn{2}{l!{\vrule width1.2bp}}{\shortstack [l] {The fraction of positive rate over false positives.}} \\
\hline
FPPI & \shortstack [c] {FP Per Image} & \multicolumn{2}{l!{\vrule width1.2bp}}{\shortstack [l] {The fraction of false positive for each image.}} \\
\hline
MR & \shortstack [c] {log-average \\missing rate} & \multicolumn{2}{l!{\vrule width1.2bp}}{\shortstack [l] {Average miss rate over different FPPI rates evenly spaced in log-space}} \\
\hline
\multicolumn{4}{!{\vrule width1.2bp}c!{\vrule width1.2bp}}{{ \bf Generic Object Detection}}    \\
\hline
\multirow{8}{*}{ mAP} & \multirow{9}{*}{\shortstack [c] {mean \\Average\\Precision}}  & VOC2007 & mAP at 0.50 IoU threshold over all 20 classes. \\
\cline{3-4}
& & VOC2012 & mAP at 0.50 IoU threshold over all 20 classes. \\
\cline{3-4}
& & OpenImages & mAP at 0.50 IoU threshold over 500 most frequent classes. \\
\cline{3-4}
& & \multirow{6}{*}{ MSCOCO}& $\bullet$ AP$_\textrm{coco}$: mAP averaged over ten $\Omega$: $\{0.5:0.05:0.95\}$;\\
& & & $\bullet$ AP$_{50}$: mAP at 0.50 IoU threshold;\\
& & & $\bullet$ AP$_{75}$: mAP at 0.75 IoU threshold;\\
& & & $\bullet$ AP$_{\it S}$: AP$_\textrm{coco}$ for small objects of area smaller than $32^2$;\\
& & & $\bullet$ AP$_{\it M}$: AP$_\textrm{coco}$ for objects of area between $32^2$ and $96^2$;\\
& & & $\bullet$ AP$_{\it L}$: AP$_\textrm{coco}$ for large objects of area bigger than $96^2$; \\
\Xhline{1pt}
\multicolumn{4}{!{\vrule width1.2bp}c!{\vrule width1.2bp}}{{\bf Face Detection}}    \\
\hline
\multirow{5}{*}{ mAP} & \multirow{5}{*}{\shortstack [c] {mean \\Average\\Precision}}  & Pascal Face & mAP at 0.50 IoU threshold. \\
\cline{3-4}
& & AFW & mAP at 0.50 IoU threshold. \\
\cline{3-4}
& & \multirow{3}{*}{WIDER FACE} &  $\bullet$ mAP$_\textrm{easy}$: mAP for easy level faces; \\
& &  &  $\bullet$ mAP$_\textrm{mid}$: mAP for mid level faces; \\
& &  &  $\bullet$ mAP$_\textrm{hard}$: mAP for hard level faces; \\
\hline
\multirow{2}{*}{ TPR} & \multirow{2}{*}{\shortstack [c] {True \\ Positive\\ Rate}}  & \multirow{2}{*}{FDDB} & $\bullet$ TPR$_{\textrm dis}$ with 1k FP at 0.50 IoU threshold, with bbox level. \\
&  & & $\bullet$ TPR$_{\textrm cont}$ with 1k FP at 0.50 IoU threshold, with eclipse level. \\
\Xhline{1pt}
\multicolumn{4}{!{\vrule width1.2bp}c!{\vrule width1.2bp}}{{ \bf Pedestrian Detection}}    \\
\hline
\multirow{3}{*}{ mAP} & \multirow{3}{*}{\shortstack [c] {mean \\Average\\Precision}}  & \multirow{3}{*}{KITTI} &$\bullet$ mAP$_\textrm{easy}$: mAP for easy level pedestrians; \\
& &  &  $\bullet$ mAP$_\textrm{mid}$: mAP for mid level pedestrians; \\
& &  &  $\bullet$ mAP$_\textrm{hard}$: mAP for hard level pedestrians; \\
\hline \multirow{5}{*}{ MR} & \multirow{5}{*}{\shortstack [c] {log-average \\ miss rate}}  & CityPersons & MR: ranging from $1e^{-2}$ to 100 FPPI \\
\cline{3-4}
& & Caltech & MR: ranging from $1e^{-2}$ to 100 FPPI \\
\cline{3-4}
& & ETH &  MR: ranging from $1e^{-2}$ to 100 FPPI \\
\cline{3-4}
& & INRIA &  MR: ranging from $1e^{-2}$ to 100 FPPI\\
\Xhline{1pt}
\end{tabular}
}
\caption {Summary of common evaluation metrics for various detection tasks including generic object detection, face detection and pedestrian detection. }\label{tab:eval}
\end{table}

\begin{table*}[hpt]
\centering
\begin{threeparttable}
\begin{tabular}{l|c|c|c|c|c}
\toprule[1.5pt]
\multirow{2}{*}{Method} &\multirow{2}{*}{Backbone} &\multirow{2}{*}{Proposed Year}&\multirow{2}{*}{Input size(Test)} &\multicolumn{2}{c}{mAP (\%)} \\
\cline{5-6}
& & && VOC2007 & VOC2012\\
\hline
\textit{Two-stage Detectors:} & & & & &\\
R-CNN \cite{girshick2014rich}   &VGG-16&2014 &Arbitrary  &66.0\tnote{$\bm{\ast}$} &62.4\tnote{$\bm{\dagger}$}\\
SPP-net \cite{girshick2014rich}   &VGG-16 &2014&$\sim600\times1000$  &63.1\tnote{$\bm{\ast}$} &-\\
Fast R-CNN \cite{girshick2015fast}   &VGG-16 &2015&$\sim600\times1000$  &70.0 &68.4\\
Faster R-CNN \cite{ren2015faster} &VGG-16 &2015&$\sim600\times1000$  &73.2 & 70.4 \\
MR-CNN \cite{gidaris2015object} &VGG-16 &2015&Multi-Scale  &78.2 & 73.9 \\
Faster R-CNN \cite{he2016deep} &ResNet-101 &2016&$\sim600\times1000$  &76.4 & 73.8 \\
R-FCN \cite{dai2016r}      &ResNet-101 &2016&$\sim600\times1000$  &80.5 & 77.6 \\
OHEM \cite{shrivastava2016training}   &VGG-16 &2016&$\sim600\times1000$  & 74.6 & 71.9 \\
HyperNet \cite{kong2016hypernet}    &VGG-16 &2016&$\sim600\times1000$  &76.3 & 71.4 \\
ION \cite{bell2016inside}       &VGG-16 &2016&$\sim600\times1000$  &79.2 & 76.4 \\
CRAFT \cite{yang2016craft}       &VGG-16 &2016&$\sim600\times1000$  &75.7 & 71.3\tnote{$\bm{\dagger}$} \\
LocNet \cite{gidaris2016locnet}       &VGG-16 &2016&$\sim600\times1000$  &78.4 & 74.8\tnote{$\bm{\dagger}$} \\
R-FCN w DCN~\cite{dai2017deformable}      &ResNet-101 &2017&$\sim600\times1000$  &82.6 & - \\
CoupleNet \cite{zhu2017couplenet} &ResNet-101 &2017&$\sim600\times1000$  &82.7 & 80.4 \\
DeNet512(wide)~\cite{tychsen2017denet} &ResNet-101 &2017&$\sim512\times512$  &77.1 & 73.9 \\
FPN-Reconfig~\cite{kong2018deep}  &ResNet-101 &2018&$\sim600\times1000$  &82.4 & 81.1 \\
DeepRegionLet~\cite{xu2018deep} &ResNet-101 &2018&$\sim600\times1000$  &83.3 & \textbf{81.3}\\
DCN+R-CNN~\cite{cheng2018revisiting} &ResNet-101+ResNet-152&2018 & Arbitrary  &\textbf{84.0} & 81.2

\\
 \hline \hline
\textit{One-stage Detectors:} & && & &\\
YOLOv1 \cite{redmon2016you}   &VGG16 &2016&$448\times448$ &66.4 & 57.9 \\
SSD512 \cite{liu2016ssd} &VGG-16 &2016&$512\times512$  &79.8 & 78.5 \\
YOLOv2 \cite{redmon2017yolo9000}  &Darknet &2017&$544\times544$  &78.6 & 73.5 \\
DSSD513 \cite{fu2017dssd} &ResNet-101 &2017&$513\times513$  &81.5 & 80.0 \\
DSOD300 \cite{shen2017dsod} &DS/64-192-48-1 &2017&$300\times300$  &77.7 & 76.3 \\
RON384 \cite{kong2017ron}   &VGG-16&2017 &$384\times384$  &75.4 & 73.0 \\
STDN513~\cite{zhou2018scale} &DenseNet-169 &2018 &$513\times513$  &80.9 & - \\
RefineDet512 \cite{zhang2018single}  &VGG-16 &2018   &$512\times512$  & 81.8 & 80.1\\
RFBNet512~\cite{liu2018receptive} &VGG16 &2018&$512\times512$  &\textbf{82.2} & - \\
CenterNet~\cite{zhou2019objects}& ResNet101&2019 &$512\times512$ &78.7&-\\
CenterNet~\cite{zhou2019objects}& DLA~\cite{zhou2019objects}&2019 &$512\times512$ &80.7&-\\

\bottomrule[1.5pt]
\end{tabular}
\begin{tablenotes}
\item[$\bm{\ast}$] This entry reports the the model is trained with VOC2007 {\tt trainval} sets only.
\item[$\bm{\dagger}$] This entry reports the the model are trained with VOC2012 {\tt trainval} sets only .
\end{tablenotes}
\caption{Detection results on PASCAL VOC dataset. For VOC2007, the models are trained on VOC2007 and VOC2012 {\tt
trainval} sets and tested on VOC2007 {\tt test} set. For VOC2012, the models are trained on VOC2007 and VOC2012 {\tt trainval}
sets plus VOC2007 {\tt test} set and tested on VOC2012 {\tt test} set by default. Since Pascal VOC datasets are well tuned and thus the number of detection frameworks for VOC reduces in recent years.}\label{tab:voc}
\end{threeparttable}
\end{table*}

\begin{table*}[hpt]
\centering

\begin{tabular}{l|c|c|ccc|cccc}
\toprule[1.5pt]
Method &Backbone&Year &~~~AP~~~ &~AP$_{50}$~ & ~AP$_{75}$~ & ~AP$_{\it S}~$ & ~AP$_{\it M}$~ & ~AP$_{\it L}$~\\
\hline
\textbf{Two-Stage Detectors:} & & & & & & & & & \\
Fast R-CNN \cite{girshick2015fast} &VGG-16 &2015 &19.7 &35.9 &- &- &- &- \\
Faster R-CNN \cite{ren2015faster} &VGG-16&2015 &21.9 &42.7 &- &- &- &- \\
OHEM \cite{shrivastava2016training} &VGG-16&2016 &22.6 &42.5 &22.2 &5.0 &23.7 &37.9 \\
ION \cite{bell2016inside} &VGG-16&2016 &23.6 &43.2 &23.6 &6.4 &24.1 &38.3\\
OHEM++ \cite{shrivastava2016training} &VGG-16&2016 &25.5 &45.9 &26.1 &7.4 &27.7 &40.3 \\
R-FCN \cite{dai2016r} &ResNet-101&2016 &29.9 &51.9 &- &10.8 &32.8 &45.0\\
Faster R-CNN+++ \cite{he2016deep} &ResNet-101 &2016&34.9 &55.7 &37.4 &15.6 &38.7 &50.9\\
Faster R-CNN w FPN \cite{lin2016fpn} &ResNet-101&2016 &36.2 &59.1 &39.0 &18.2 &39.0 &48.2 \\

DeNet-101(wide)~\cite{tychsen2017denet} & ResNet-101&2017 &33.8 &53.4 &36.1 &12.3 &36.1 &50.8 \\
CoupleNet \cite{zhu2017couplenet} &ResNet-101&2017 &34.4 &54.8 &37.2 &13.4 &38.1 &50.8 \\
Faster R-CNN by G-RMI \cite{huang2017speed} &Inception-ResNet-v2&2017 &34.7 &55.5 &36.7 &13.5 &38.1 &52.0 \\
Deformable R-FCN \cite{dai2016r} &Aligned-Inception-ResNet&2017 &37.5 &58.0 &40.8 &19.4 &40.1 &52.5 \\
Mask-RCNN~\cite{he2017mask} & ResNeXt-101&2017 &39.8 &62.3 &43.4 &22.1 &43.2 &51.2  \\
umd\_det \cite{bodla2017improving}             &ResNet-101&2017   &40.8 &62.4 &44.9 &23.0 &43.4 &53.2 \\
Fitness-NMS~\cite{tychsen2017improving}     &ResNet-101&2017 &41.8 &60.9 &44.9 &21.5 &45.0 &57.5\\

DCN w Relation Net~\cite{hu2018relation}     &ResNet-101&2018 &39.0 &58.6 &42.9 &- &- &-\\
DeepRegionlets~\cite{xu2018deep}    &ResNet-101&2018 &39.3 &59.8 &- &21.7 &43.7 &50.9\\
C-Mask RCNN~\cite{chen2018context}    &ResNet-101&2018 &42.0 &62.9 &46.4 &23.4 &44.7 &53.8\\
Group Norm~\cite{wu2018group}     &ResNet-101&2018 &42.3 &62.8 &46.2 &- &- &-\\
DCN+R-CNN~\cite{cheng2018revisiting} &ResNet-101+ResNet-152&2018 &42.6 &65.3 &46.5 &26.4 &46.1 &56.4\\
Cascade R-CNN \cite{cai2018cascade} & ResNet-101&2018 & 42.8 & 62.1 & 46.3 &23.7 &45.5 &55.2  \\
SNIP++~\cite{singh2018analysis} &DPN-98&2018 &45.7& 67.3 &51.1 &29.3 &48.8 &57.1\\
SNIPER++~\cite{singh2018sniper}  &ResNet-101&2018 &46.1& 67.0 &51.6 &29.6 &48.9 &58.1\\
PANet++~\cite{liu2018path}& ResNeXt-101&2018& 47.4 &67.2 &51.8& 30.1 &51.7 &60.0\\
Grid R-CNN~\cite{lu2019grid}     &ResNeXt-101&2019 &43.2 &63.0 &46.6 &25.1 &46.5 &55.2\\
DCN-v2~\cite{zhu2019deformv2}     &ResNet-101&2019 &44.8 &66.3 &48.8 &24.4 &48.1 &59.6\\
DCN-v2++~\cite{zhu2019deformv2}     &ResNet-101&2019 &46.0 &67.9 &50.8 &27.8 &49.1 &59.5\\
TridentNet~\cite{li2019scale} &ResNet-101&2019 &42.7 &63.6& 46.5& 23.9& 46.6& 56.6\\
TridentNet~\cite{li2019scale} &ResNet-101-Deformable&2019& \bf{48.4}& \bf{69.7}& \bf{53.5}& 31.8& 51.3& 60.3\\
\hline
\hline
\textbf{Single-Stage Detectors:} & & & & & & & & \\
SSD512 \cite{liu2016ssd} &VGG-16&2016 &28.8 &48.5 &30.3 &10.9 &31.8 &43.5\\
RON384++ \cite{kong2017ron} &VGG-16&2017 &27.4 &49.5 &27.1 &- &- &- \\

YOLOv2 \cite{redmon2017yolo9000} &DarkNet-19&2017 &21.6 &44.0 &19.2 &5.0 &22.4 &35.5\\
SSD513 \cite{fu2017dssd} &ResNet-101&2017 &31.2 &50.4 &33.3 &10.2 &34.5 &49.8 \\
DSSD513 \cite{fu2017dssd} &ResNet-101&2017 &33.2 &53.3 &35.2 &13.0 &35.4 &51.1 \\
RetinaNet800++ \cite{lin2017focal} &ResNet-101&2017 &39.1 &59.1 &42.3 &21.8 &42.7 &50.2 \\

STDN513~\cite{zhou2018scale}& DenseNet-169&2018 &31.8 &51.0 &33.6 &14.4 &36.1 &43.4 \\
FPN-Reconfig \cite{kong2018deep} &ResNet-101&2018 &34.6 &54.3 &37.3 &- &- &- \\
RefineDet512 \cite{zhang2018single}  &ResNet-101&2018 &36.4 &57.5 &39.5 &16.6 &39.9 &51.4 \\
RefineDet512++ \cite{zhang2018single}  &ResNet-101&2018 &41.8 &62.9 &45.7 &25.6 &45.1 &54.1 \\
GHM SSD~\cite{li2018gradient} & ResNeXt-101&2018 &41.6 &62.8 &44.2 &22.3 &45.1 &55.3 \\
CornerNet511~\cite{law2018cornernet} & Hourglass-104&2018 & 40.5 &56.5 &43.1 &19.4 &42.7 &53.9 \\
CornerNet511++~\cite{law2018cornernet} & Hourglass-104&2018 & 42.1 &57.8 & 45.3 &20.8 &44.8 & 56.7\\
M2Det800~\cite{zhao2019m2det}& VGG-16&2019 & 41.0 &59.7 &45.0 &22.1 &46.5 &53.8 \\
M2Det800++~\cite{zhao2019m2det}& VGG-16&2019 & 44.2 &64.6 &49.3 &29.2 &47.9 &55.1\\
ExtremeNet~\cite{zhou2019bottom} & Hourglass-104 &2019&40.2 & 55.5 &43.2 & 20.4 & 43.2 & 53.1 \\
CenterNet-HG~\cite{zhou2019objects} &Hourglass-104&2019 &42.1 & 61.1 &45.9& 24.1& 45.5& 52.8 \\
FCOS~\cite{tian2019fcos}& ResNeXt-101&2019& 42.1& 62.1& 45.2& 25.6& 44.9& 52.0\\
FSAF~\cite{zhu2019feature} &ResNeXt-101&2019 &42.9& 63.8& 46.3& 26.6& 46.2& 52.7\\ 
CenterNet511~\cite{duan2019centernet} & Hourglass-104&2019& 44.9& 62.4& 48.1& 25.6& 47.4& 57.4\\
CenterNet511++~\cite{duan2019centernet}& Hourglass-104&2019 & \bf{47.0} &\bf{64.5}& \bf{50.7} &28.9& 49.9& 58.9\\

\bottomrule[1.5pt]
\end{tabular}
\caption{Detection performance on the MS COCO {\tt test-dev} data set. "++" denotes applying inference strategy such as multi scale test, horizontal flip, etc.}
\label{tab:coco}
\end{table*}

\subsection{Face Detection Benchmarks}
In this section, we introduce several widely used face detection datasets (WIDER FACE, AFW, FDDB and Pascal Face) and the commonly used evaluation metrics. 

\noindent{\bf WIDER FACE~\cite{yang2016wider}.} WIDER FACE has totally 32203 images with about 400k faces for a large range of scales. It has three subsets: 40\% for training, 10\% for validation, and 50\% for test. The annotations of training and validation sets are online available. According to the difficulty of detection tasks, it has three splits: Easy, Medium and Hard.

\noindent{\bf FDDB~\cite{fddbTech}.} The Face Detection Data set and Benchmark (FDDB) is a well-known benchmark with 5171 faces in 2845 images. Commonly face detectors will first be trained on a large scale dataset(WIDERFACE etc. ) and tested on FDDB.

\noindent{\bf PASCAL FACE~ \cite{Everingham10}.} This dataset was collected from PASCAL person layout test set, with 1335 labeled faces in 851 images. Similar to FDDB, it's commonly used as test set only.

\noindent{\bf Evaluation Metrics.} As Tab. \ref{tab:eval} shown, the evaluation metric for WIDER FACE and PASCAL FACE is mean average precision (mAP) with IoU threshold as 0.5, and for WIDER FACE the results of each difficulty level will be reported. For FDDB, true positive rate (TPR) at 1k false positives are used for evaluation. There are two annotation types to evaluate FDDB dataset: bounding box level and eclipse level.

\subsection{Pedestrian Detection Benchmarks}
In this section we will first introduce five widely used datasets(Caltech, ETH, INRIA, CityPersons and KITTI) for pedestrian object detection and then introduce their evaluation metrics.

\noindent{\bf CityPersons~\cite{zhang2017citypersons}} is a new pedestrian detection dataset on top of the semantic segmentation dataset CityScapes~\cite{Cordts2016Cityscapes}, of which 5000 images are captured in several cities in Germany. A total of 35000 persons with an additional 13000 ignored regions, both bounding box annotation of all persons and annotation of visible parts are provided.

\noindent{\bf Caltech~\cite{dollar2012pedestrian} } is one of the most popular and challenging datasets for pedestrian detection, which comes from approximately 10 hours 30Hz VGA video recorded by a car traversing the streets in the greater Los Angeles metropolitan area. The training and testing sets contains 42782 and 4024 frames, respectively.

\noindent{\bf ETH~\cite{ess2007depth}} contains 1804 frames in three video clips and commonly it's used as test set to evaluate performance of the models trained on the large scale datasets(CityPersons dataset etc.).

\noindent{\bf INRIA~\cite{dalal2005histograms}} contains images of high resolution pedestrians collected mostly from holiday photos, which consists of 2120 images, including 1832 images for training and 288 images. Specifically, there are 614 positive images and 1218 negative images in the training set.

\noindent{\bf KITTI~\cite{Geiger2013IJRR}} contains 7481 labeled images of resolution 1250x375 and another 7518 images for testing. The person class in KITTI is divided into two subclasses: pedestrian and cyclist, both evaluated by mAP method. KITTI contains three evaluation metrics: easy, moderate and hard, with difference in the min. bounding box height, max. occlusion level, etc.

\noindent{\bf Evaluation Metrics.} For CityPersons, INRIA and ETH, the log-average miss rate over 9 points ranging from $1e^{-2}$ to 100 FPPI(False Positive Per Image) is used to evaluate the performance of the detectors(lower is better). For KITTI, standard mean average precision is used as evaluation metric with 0.5 IoU threshold.

\section{State-of-the-art for Generic Object Detection}\label{sec:sota}
Pascal VOC2007, VOC2007 and MSCOCO are three most commonly used datasets for evaluating detection algorithms. Pascal VOC2012 and VOC2007 are mid scale datasets with 2 or 3 objects per image and the range of object size in VOC dataset is not large. For MSCOCO, there are nearly 10 objects per image and the majority objects are small objects with
large scale ranges, which leads to a very challenge task for detection algorithms. In Table \ref{tab:coco} and Table \ref{tab:voc} we give the benchmarks of VOC2007, VOC2012 and MSCOCO over the recent few years.

\section{Concluding Remarks and Future Directions}\label{sec:conc}

Object detection has been actively investigated and new state-of-the-art results have been reported almost every few months. However, there are still many open challenges. Below we discuss sveral open challenges and future directions.  

\noindent{\bf i) Scalable Proposal Generation Strategy.} As claimed in Sec. \ref{sec:prop}, currently most detectors are anchor-based methods, and there are some critical shortcomings which limit the detection accuracy. Current anchor priors are mainly manually designed which is difficult to match multi-scale objects and the matching strategy based on IoU is also heuristic. Although some methods have been proposed to transform anchor-based methods into anchor-free methods (e.g. methods based on keypoints), there are still some limitations(high computation cost etc.) with large space to improve. From Figure \ref{fig:milestone}, developing anchor-free methods becomes a very hot topic in object detection~\cite{law2018cornernet,zhu2019feature,zhou2019bottom,tian2019fcos,duan2019centernet}, and thus designing an efficient and effective proposal generation strategy is potentially a very important research direction in the future. 

\noindent{\bf ii) Effective Encoding of Contextual Information.} Contexts can contribute or impede visual object detection results, as objects in the visual world have strong relationships, and contexts are critical to better understand the visual worlds. However, little effort has been focused on how to correctly use contextual information. How to incorporate contexts for object detection effectively can be a promising future direction. 

\noindent{\bf iii) Detection based on Auto Machine Learning(AutoML).} To design an optimal backbone architecture for a certain task can significantly improve the results but also requires huge engineering effort. Thus to learn backbone architecture directly on the datasets is a very interesting and important research direction. From Figure \ref{fig:milestone}, inspired by the pioneering AutoML work on image classification~\cite{zoph2018learning,tan2019efficientnet}, more relevant work has been proposed to address detection problems via AutoML~\cite{chen2019detnas,ghiasi2019fpn}, such as learning FPN structure~\cite{ghiasi2019fpn} and learning data augmentation policies~\cite{zoph2019learning}, which show significant improvement over the baselines. However, the required computation resource for AutoML is unaffordable to most researchers(more than 100 GPU cards to train a single model). Thus, developing a low-computation framework shall have a large impact for object detection. Further, new structure policies (such as proposal generation and region encoding) of detection task can be explored in the future.

\noindent{\bf iv) Emerging Benchmarks for Object Detection.} Currently MSCOCO is the most commonly used detection benchmark testbed. However, MSCOCO has only 80 categories, which is still too small to understand more complicated scenes in real world. Recently, a new benchmark dataset  LVIS~\cite{Gupta_2019_CVPR} has been proposed in order to collect richer categorical information.  LVIS contains 164000 images with 1000+ categories, and there are total of 2.2 million high-quality instance segmentation masks. Further, LVIS simulates the real-world low-shot scenario where a large number of categories are present but per-category data is sometimes scarce. LVIS will open a new benchmark for more challenging detection, segmentation and low-shot learning tasks in near future. 

\noindent{\bf v) Low-shot Object Detection.} Training detectors with limited labeled data is dubbed as Low-shot detection. Deep learning based detectors often have huge amount of parameters and thus are data-hungry, which require large amount of labeled data to achieve satisfactory performance. However, labeling objects in images with bounding box level annotation is very time-consuming. Low-shot learning has been actively studied for classification tasks, but only a few studies are focused on detection tasks. For example, Multi-modal Self-Paced Learning for Detection (MSPLD)~\cite{dong2017few} addresses the low-shot detection problem in a semi-supervised learning setting where a large-scale unlabeled dataset is available. 
RepMet~\cite{schwartz2019repmet} adopts a Deep Metric Learning (DML) structure, which jointly learns feature embedding space and data distribution of training set categories. However, RepMet was only tested on datasets with similar concepts (animals). Low-Shot Transfer Detector (LSTD)~\cite{chen2018lstd} addresses low-shot detection based on transfer learning which transfers the knowledge form large annotated external datasets to the target set by knowledge regularization. LSTD still suffers from overfitting. There is still a large room to improve the low-shot detection tasks. 

\noindent{\bf vi) Backbone Architecture for Detection Task.} It has become a paradigm to adopt weights of classification models pre-trained on large scale dataset into detection problem. However, there still exist conflicts between classification and detection tasks~\cite{li2018detnet}, and thus it's not an optimal solution to do so. From Table \ref{tab:coco}, most state-of-the-art detection algorithms are based on classification backbones, and only a few of them try different selections (such as CornerNet based on Hourglass Net). Thus how to develop a detection-aware backbone architecture is also an important research direction in the future.

\noindent{\bf vii) Other Research Issues.} In addition, there are some other open research issues, such as large batch learning~\cite{peng2018megdet} and incremental learning~\cite{shmelkov2017incremental}. Batch size is a key factor in DCNN training but has not been well studied in detection task. And for incremental learning, detection algorithms still suffer from catastrophic forgetting if adapted to a new task without initial training data. These open and fundamental research issues also deserve more attention for future work.

In this survey, we give a comprehensive survey of recent advances in deep learning techniques for object detection tasks. The main contents of this survey are divided into three major categories: object detector components, machine learning strategies, real-world applications and benchmark evaluations. We have reviewed a large body of representative articles in recent literature, and presented the contributions on this important topic in a structured and systematic manner. We hope this survey can give readers a comprehensive understanding of object detection with deep learning and potentially spur more research work on object detection techniques and their applications. 

\section*{References}

\bibliography{DissertationBib}

\end{document}